\documentclass[twoside,11pt]{article}

\usepackage{blindtext}

\usepackage[preprint]{jmlr2e}
\usepackage{tabularx}
\usepackage{graphicx}%
\usepackage{multirow}%
\usepackage{amsmath,amssymb,amsfonts}%
\usepackage{mathrsfs}%
\usepackage[title]{appendix}%
\usepackage{xcolor}%
\usepackage{textcomp}%
\usepackage{manyfoot}%
\usepackage{booktabs}%
\usepackage{algorithm}%
\usepackage{algorithmicx}%
\usepackage{algpseudocode}%
\usepackage{listings}%
\usepackage[utf8]{inputenc}
\usepackage{float}
\usepackage{lineno}
\usepackage{comment}
\usepackage{pdflscape}
\usepackage{tikz}
\usepackage{multirow}
\usepackage{geometry}
\usepackage{array}
\usepackage{booktabs}
\usepackage{tabu}
\usepackage{booktabs} 
\usepackage{tabularray}

\newmuskip\pFqmuskip

\newcommand*\pFq[6][8]{%
  \begingroup % only local assignments
  \pFqmuskip=#1mu\relax
  \mathchardef\normalcomma=\mathcode`,
  \mathcode`\,=\string"8000
  \begingroup\lccode`\~=`\,
  \lowercase{\endgroup\let~}\pFqcomma
  {}_{#2}F_{#3}{\left(\genfrac..{0pt}{}{#4}{#5};#6\right)}%
  \endgroup
}
\newcommand{\pFqcomma}{{\normalcomma}\mskip\pFqmuskip}

\newmuskip\Gmuskip
\newcommand*\G[6][8]{%
  \begingroup % only local assignments
  \Gmuskip=#1mu\relax
  \mathchardef\normalcomma=\mathcode`,
  \mathcode`\,=\string"8000
  \begingroup\lccode`\~=`\,
  \lowercase{\endgroup\let~}\Gcomma
  G^{#2}_{#3}{\left(#6\middle\vert\genfrac..{0pt}{}{#4}{#5}\right)}%
  \endgroup
}
\newcommand{\Gcomma}{{\normalcomma}\mskip\Gmuskip}

\newmuskip\Fmuskip
\newcommand*\F[7][8]{%
  \begingroup % only local assignments
  \Fmuskip=#1mu\relax
  \mathchardef\normalcomma=\mathcode`,
  \mathcode`=\string"8000
  \begingroup\lccode`\~=`\,
  \lowercase{\endgroup\let~}\Fcomma
  F^{#2}_{#3}{\left(\genfrac..{0pt}{}{#4}{#5};#6,#7\right)}%
  \endgroup
}
\newcommand{\Fcomma}{{\normalcomma}\mskip\Fmuskip}

\def\R{\mathbb{R}}

\def\b_eta{\boldsymbol{\eta}}

\def\bx{\boldssymbol{x}}

\def\bx{\boldsymbol{x}}

\def\bm{\boldsymbol}

\def\R{\mathbb{R}}
\def\b_eta{\boldsymbol{\eta}}

\def\calH{\cal H}  %RS
\def\bx{\boldsymbol{x}} %RS
\def\bm{\boldsymbol}
\def\btheta{\boldsymbol{\theta}}

\usepackage{lastpage}
\jmlrheading{XX}{2025}{1-\pageref{LastPage}}{1/25; Revised }{}{21-0000}{Xavier Emery, Emilio Porcu and Moreno Bevilacqua}

\ShortHeadings{Unified Native Spaces in Kernel Methods}{Emery, Porcu and Bevilacqua}
\firstpageno{1}

\begin{document}

\title{Unified Native Spaces in Kernel Methods}

\author{\name Xavier Emery \email xemery@ing.uchile.cl \\
       \addr Department of Mining Engineering, Universidad de Chile \\
       Santiago 8370448, Chile\\
       \& Advanced Mining Technology Center, Universidad de Chile\\
       Santiago 8370448, Chile
       \AND
       \name Emilio Porcu \email emilio.porcu@ku.ac.ae \\
       \addr Department of Mathematics, Khalifa University\\
       Abu Dhabi 127788, United Arab Emirates\\
       \& ADIA Lab\\
       Abu Dhabi 127788, United Arab Emirates
        \AND
       \name Moreno Bevilacqua \email moreno.bevilacqua@uai.cl \\
       \addr Facultad de Ingenieria y Ciencias, Universidad Adolfo Ibañez\\
       Viña del Mar 2580335, Chile\\
       \& Dipartimento di Scienze Ambientali, Informatica e Statistica, Ca’ Foscari University of Venice\\
       Venice 30123, Italy}

\editor{Brian Kulis}

\maketitle

\begin{abstract}%   <- trailing '%' for backward compatibility of .sty file
There exists a plethora of parametric models for positive definite kernels, and their use is ubiquitous in disciplines as diverse as statistics, machine learning, numerical analysis, and approximation theory. Usually, the kernel parameters index certain features of an associated process. Amongst those features, smoothness (in the sense of Sobolev spaces, mean square differentiability, and fractal dimensions), compact or global supports, and negative dependencies (hole effects) are of interest to several theoretical and applied disciplines. 
This paper unifies a wealth of well-known kernels into a single parametric class that encompasses them as special cases, attained either by exact parameterization or through parametric asymptotics. 
We furthermore characterize the Sobolev space that is norm equivalent to the RKHS associated with the new kernel. As a by-product, we infer the Sobolev spaces that are associated with existing classes of kernels.  
We illustrate the main properties of the new class, show how this class can switch from compact to global supports, and provide special cases for which the kernel attains negative values over nontrivial intervals. Hence, the proposed class of kernel is the reproducing kernel of a very rich Hilbert space that contains many special cases, including the celebrated Mat\'ern and Wendland kernels, as well as their aliases with hole effects.
\end{abstract}


\section{Introduction}

The terminology {\em Native Spaces} for reproducing kernel Hilbert spaces (RKHSs) associated with given classes of positive definite kernels was originally adopted by \cite{schaback1995creating}. Positive definite kernels and native spaces are now used in approximation theory, numerical analysis, computational science, signal processing, machine learning, statistics and probability, with a monumental literature from all these disciplines and with countless applications in science and engineering. In this paper, one of the most important aspects is provided by the connection between certain parametric classes of kernels and their native spaces that are norm equivalent to certain classes of Sobolev spaces. This aspect puts {\em smoothness} into play, and smoothness plays a dominant role in the aforementioned disciplines, among others.  

\subsection{Features of Interest}

\textbf{Smoothness.} The local behavior of a stationary Gaussian random field depends on its covariance kernel; in particular, the sample paths are $k$-times differentiable in the mean-square sense if, and only if, the kernel is $2k$-times differentiable at the origin \citep{Chiles2012}. Under the same condition, the sample paths have (local) Sobolev space exponent being identically equal to $k$. While the well-known Mat\'ern kernels \citep{Matern1960} generate Sobolev spaces directly, other classes of kernels allow for RKHSs that are norm equivalent to certain classes of Sobolev spaces, in particular the Generalized Wendland kernel \citep[and the references therein]{BFFP} that parameterizes smoothness in a similar fashion as the Mat\'ern kernel. \\\\
\textbf{Support.} Compactly supported kernels play an important role in several disciplines. To mention: 
\begin{enumerate}
\item Compact support implies sparse covariance matrices, i.e., matrices with most of the elements being zero, which considerably reduces the computational burden to solve linear systems of equations %\citep{Scott2023} 
needed in spatial statistics for maximum likelihood estimations or for prediction by kriging \citep{furrer2006covariance, sun2012geostatistics}.
\item The discrete Fourier spectra of the covariance matrix computed on a sufficiently large spatial domain has nonnegative entries, which allows the exact simulation of Gaussian random fields on regular grids via the discrete spectral approach \citep[Section 7.5.4]{Chiles2012}. Without the compact support restriction, the simulation becomes approximate and does not accurately reproduce the spatial correlation structure of the target random field. 
\item Isotropic kernels on the $d$-dimensional unit sphere can be constructed from isotropic kernels in the $d$-dimensional Euclidean space $\mathbb{R}^d$, which can be done by substituting the Euclidean distance by the geodesic distance on the sphere, provided that $d$ is odd and that the kernel is supported in $[0,\pi]$ \citep{emery2023CA}.
\item Transitive covariograms and geometric covariograms are compactly-supported kernels used in geostatistics, mathematical morphology, and stochastic geometry \citep{matheron1965, serra1982}%,stoyan}.
\end{enumerate}
\textbf{Hole effects.} An isotropic kernel in $\R^d$ are lower-bounded by $-\frac{1}{d}$ \citep{Matern1960}. The Mat\'ern kernel is valid in any dimension, and this explains why it only attains strictly positive values. This is certainly a limitation. Kernels attaining negative values are said to have a {\em hole effect} in geostatistics \citep{Chiles2012}. Such a feature is of interest in various disciplines of the natural sciences and engineering, and the reader is referred to the online supplement in \cite{alegria2023} for a complete account.

\subsection{Challenges and Contribution}

To the best of the authors' knowledge, the three aforementioned features (smoothness, compact support, and hole effects) have hardly been simultaneously considered in a single class of positive definite kernels. Substantially, the literature has challenged the smoothness problem using the Mat\'ern class, and the computational problem using the compactly supported Generalized Wendland class. As a result, the literature is fragmented, with a clear lack of connections. 
Additionally, many radially symmetric kernels proposed in earlier literature have not been properly characterized in terms of smoothness. 

This paper presents a new class of radially symmetric positive definite kernels that includes most of the parametric classes of kernels from previous literature as special cases. These cases can be attained either through specific parameterization, or through asymptotics on the parameters indexing the new class. 
As a result, this new class encompasses kernels with either compact of global support, and kernels being strictly positive or exhibiting a hole effect. 

We furthermore find the Sobolev space that is norm equivalent to the RKHS associated with this new class. As a by-product, all the well-known classes of positive definite kernels are implicitly characterized in terms of smoothness. Moreover, we infer the local properties of the proposed kernels in terms of ordinary $2k$-order derivatives, which in turn determines the $k$-mean square differentiability of associated Gaussian random fields. Similarly, a study of the local properties of the proposed kernel (through elegant Tauberian inversions) allows to determine the fractal dimension \citep{falconer2014fractal} of the associated Gaussian fields. As a by-product, we provide a characterization of these properties for the majority of kernels proposed in earlier literature. 

\subsection{How to Read this Paper}

Readers who are not familiar with mathematics should not disregard messages contained in this paper. A look at Table \ref{tab:hygeo} and Figure \ref{figure_relations} (see Section \ref{sec: construction} next) will provide an immediate impact of this paper. A wealth of well-known kernels, except those with heavy tails that are beyond the scope of this paper, are included as special or asymptotic cases of the new kernel and are implicitly characterized in terms of smoothness. Practitioners can now control the three main modeling features -- smoothness, compact support and hole effect -- through a single class of kernels. 

The outline of the paper is as follows. A succinct background in Section \ref{sec2} illustrates the connections between positive definite kernels, RKHSs and Sobolev spaces. Section \ref{sec: construction} provides the new parametric class of kernels and the parametric space for which the kernel is {\em permissible} (read: positive definite). Further, we characterize the spectral density related to this class. An asymptotic argument will allow to determine the Sobolev space that is norm equivalent to the native space associated with the new class. This section also provides a study of the local properties of Gaussian random fields with the new covariance kernels proposed here. The special cases indicated in Table \ref{tab:hygeo} are attained through specific parameterizations from the new class (Section \ref{sec: special}), or through parametric asymptotics (Section \ref{sec: convergence}). Section \ref{sec: consequences} illustrates some consequences of our finding for current and prospective literature in both statistics and machine learning. Concluding remarks are provided in Section \ref{sec:concl}. Readers interested in the mathematical proofs can further integrate the reading through Appendices \ref{walk} to \ref{appProofs}, which contain many results of independent interest.  

\section{Background} 
\label{sec2}

\subsection{Notation}
\begin{itemize}
\item $\mathsf{i}$ is the complex imaginary unit.
\item $\mathbb{R}_{> \alpha}$ is the set of real numbers greater than $\alpha$.
\item $\mathbb{N}_{\geq \alpha}$ is the set of integers greater than, or equal to, $\alpha$; $\mathbb{N}_{\geq 0}$ is simply denoted as $\mathbb{N}$.
\item $\langle \cdot, \cdot \rangle_d$ is the inner product in $\R^d$.
\item $\| \cdot \|_d$ is the Euclidean norm in $\R^d$.
\item $(\cdot)_+$ is the positive part function.
\item $\lfloor \cdot \rfloor$ and $\lceil \cdot \rceil$ are respectively the floor and ceil functions.
\item $(\cdot)_n$ is the Pochhammer symbol.
\item $\Gamma$ is the Gamma function.
\item $\Gamma^+(\cdot,\cdot)$ and $\Gamma^-(\cdot,\cdot)$ are the upper and lower incomplete Gamma functions, respectively.
\item $J_{\nu}$ is the Bessel function of the first kind.
\item $K_{\nu}$ is the modified Bessel function of the second kind. 
\item $L_n^{\mu}$ is the generalized Laguerre polynomial.
\item ${}_2F_1({}^{\alpha,\beta}_{\gamma};\cdot)$ is the Gauss hypergeometric function, with $\alpha, \beta,\gamma$ real. 
\item ${}_pF_q({}^{\boldsymbol{\beta}}_{\boldsymbol{\gamma}};\cdot)$ is the generalized hypergeometric function, with $p,q \in \mathbb{N}$, $\boldsymbol{\beta} \in \mathbb{R}^p, \boldsymbol{\gamma}\in \mathbb{R}^q$. 
\item $G_{p,q}^{m,n}(\cdot \big| {}^{\boldsymbol{\beta}}_{\boldsymbol{\gamma}})$ is the Meijer-$G$ function, with $p,q \in \mathbb{N}$, $\boldsymbol{\beta} \in \mathbb{R}^p$, $\boldsymbol{\gamma} \in \mathbb{R}^q$.
\end{itemize}
For an overview of special functions, see \cite{exton} and \cite{Olver}.

\subsection{Random Fields, Kernels and Native Spaces}
Let $d \in \mathbb{N}_{\geq 1}$ and $Z=\{{Z}(\bm{x}): \bm{x} \in \R^d \}$ be a Gaussian random field having zero mean and kernel (\emph{covariance function}) ${K}: \R^d \times \R^d \to \R$ defined through 
$K(\bm{x},\bm{x}^{\prime}):={\rm Cov}(Z(\bm{x}),Z(\bm{x}^{\prime}))$.
Covariance functions are symmetric and positive (semi)definite, that is
\begin{equation*} 
%\label{posdef}
\sum_{i=1}^n \sum_{j=1}^n {c}_i {K}(\bm{x}_i, \bm{x}_j) {c}_j \geq 0
\end{equation*}
for all $n \in \mathbb{N}$, $c_1, \ldots,c_n \in \mathbb{R}$ and $\bm{x}_1,\ldots,\bm{x}_n \in \mathbb{R}^d$.
If the above inequality is strict for $(c_1,\ldots,c_n)$ being non-zero, then ${K}$ is called strictly positive definite. 
Positive definite (and symmetric) functions ${K}: \R^d \times \R^d \to \R$
determine {\em translate} functions $K(\bx,\cdot)$ on $\R^d$, for all $\bx\in\R^d$. We define the inner product applying on pairs of translates through 
\begin{equation}
\label{eqHKKK}
\Big \langle  K(\bx,\cdot), K(\bm{x}^{\prime},\cdot) \Big \rangle_{\calH(K)}
:=K(\bx, \bm{x}^{\prime}), \; \; \bx,\,\bm{x}^{\prime}\in \R^d.
\end{equation}
Such an inner product extends to all linear combinations of translates and {\em generates}, by completion, a Hilbert space $\calH(K)$ of functions on $\R^d$. We follow \cite{schaback1995creating} to call this space  the {\em native} space for $K$. In most cases, this Hilbert space is a subspace of the space $L_2(\R^d)$ of continuous and square integrable functions in $\R^d$.  
As explained in \cite{schaback1995creating}, the Hilbert space allows for continuous point evaluations $\delta_{\bm x}\;:\;f\mapsto f({\bm x})$ via a {\em reproduction formula} of the type
\begin{equation*}
%\label{eqrepro}
f({\bm x})=\Big \langle f,K({\bm x},\cdot) \Big \rangle_{{\cal H}(K)},\;{\bm x}\in \R^d,\; \; f\in {\cal H}(K),
\end{equation*}
which directly follows from (\ref{eqHKKK}).
The space  ${\cal H}(K)$ is called a {\em reproducing kernel Hilbert space} with reproducing \emph{kernel} $K$. 
We note that the translates 
$K({\bm x},\cdot)$ lie in ${\cal H}(K)$, forming its completion and
being the Riesz representers of delta functionals $\delta_{\bm x}$. Translates cover a central role in numerical analysis, approximation theory and machine learning, because the so-called kernel trick allows for computing inner products over the abstract space ${\cal H}(K)$. 

Our paper deals with kernels that are stationary and isotropic (radially symmetric). Yet, it is useful to start with the assumption of stationarity only, for which the Fourier transform of $K$, denoted $\hat K$, can be used to recover the inner product (\ref{eqHKKK}) on the Hilbert space ${\cal H}(K)$ through
\begin{equation}
\label{eqHK}
    \langle f,g\rangle_{{\cal H}(K)}=\int_{\R^d}\frac{\hat f(\bm{\omega})
\overline{\hat g(\bm{\omega})}}{\hat K(\bm{\omega})}{\rm d}\bm{\omega},
\; \; f,g\in {\cal H}(K),
\end{equation}
up to a constant factor, see \cite{porcu2024matern} for details. 
Hereinafter, $\overline{g}$ denotes the complex conjugate of a function $g$, {and $\hat{g}$ its Fourier transform.}  
We can rephrase the above by saying that the space ${\cal H}(K)$ contains those functions $f$ such that the ratio $\hat f {\hat K}^{-1/2}$ is square integrable over $\mathbb{R}^d$. 
This fact creates an important link with certain function spaces. 

Under the stationarity assumption, we have  $K(\bm{x},\bm{x}^{\prime}) \equiv K(\bm{x} - \bm{x}^{\prime})$. Continuous and stationary kernels are uniquely determined as the Fourier transforms of nondecreasing, bounded and symmetric measures \citep{Yaglom:1987}:%\citep{bochner1955harmonic}: 
\begin{equation*}
K(\bm{x} - \bm{x}^{\prime}) =  \int_{\R^{d}} {\rm e}^{\mathsf{i} \langle \bm{x} - \bm{x}^{\prime} \, , \,  \bm{\omega} \rangle_d} F ({\rm d} \bm{\omega}), \qquad \bm{x} , \bm{x}^{\prime} \in \R^d.
\end{equation*}
Under some regularity conditions, Fourier inversion is feasible, and the inverse Fourier transform is called {\em spectral density} and denoted $\widehat{K}$ throughout. 

\subsection{Spectral Representations of Stationary Isotropic Kernels}

We now turn into the additional assumption of isotropy or radial symmetry, for which $K(\boldsymbol{x},\boldsymbol{x}^\prime)$ exists for any $\boldsymbol{x}$ and $\boldsymbol{x}^\prime$ in $\mathbb{R}^{d}$ and only depends on the distance $\| \boldsymbol{x}-\boldsymbol{x}^\prime \|_d$:
\begin{equation}
\label{eq:stationarycov_Rd}
K(\boldsymbol{x},\boldsymbol{x}^\prime) = {C}\left(\| \boldsymbol{x}-\boldsymbol{x}^\prime \|_d \right), \quad \boldsymbol{x},\boldsymbol{x}^{\prime} \in \mathbb{R}^{d}.
\end{equation}

The positive definiteness of $K$ implies that the matrix $[C(\| \boldsymbol{x}_i - \boldsymbol{x}_j \|_d)]_{i,j=1}^p$ is symmetric positive definite for any positive integer $p$ and any choice of $\boldsymbol{x}_1, \ldots, \boldsymbol{x}_p \in \mathbb{R}^d$. Hereinafter, we denote $\Phi_{d}$ the class of continuous mappings ${C}:[0,+\infty) \to \mathbb{R}$ such that (\ref{eq:stationarycov_Rd}) is true for a second-order stationary isotropic random field in $\mathbb{R}^{d}$. One has:
\begin{equation*}
  \Phi_1 \supset \Phi_2 \supset \ldots \supset \Phi_d \supset \ldots \supset \Phi_{\infty} := \bigcap_{d=1}^{+\infty} \Phi_d.
\end{equation*}

Any member of $\Phi_d$ has the following representation  \citep[Theorem 1]{schoenberg1938metric2}:
\begin{equation}
\label{schoenbergmeasure}
    C(h) =  \int_0^{+\infty} {\cal J}_{1/u,d}(h) \text{d}G_d(u), \qquad h \geq 0,
\end{equation}
where $G_d$ is a nondecreasing bounded measure on $(0,+\infty)$ (called Schoenberg measure by \cite{daley2014dimension}) and ${\cal J}_{a,d}$ is the Schoenberg (aka Bessel-J) kernel: 
\begin{equation}
\label{schoenbergJ}
{\cal J}_{a,d}(h) := \begin{cases}
\Gamma\left(\frac{d}{2}\right) \left(\frac{h}{2a} \right)^{1-\frac{d}{2}} J_{\frac{d}{2}-1}\left(\frac{h}{a} \right) \text{ if $h > 0$}\\
1 \text{ if $h=0$}.
\end{cases}
\end{equation}

If, furthermore, $C(\| \cdot \|_d)$ is absolutely integrable in $\R^d$, then $G_d$ is absolutely continuous with respect to the Lebesgue measure, i.e., it has a density $g_d$ such that:
\begin{equation}
    \label{schoenberg2}
    C(h) =  2^{\frac{d}{2}-1} \Gamma\left(\frac{d}{2}\right) h^{1-\frac{d}{2}} \int_0^\infty u^{1-\frac{d}{2}} J_{\frac{d}{2}-1}(uh) g_d(u) \text{d}u, \qquad h > 0,
\end{equation}
where $g_d$ is a nonnegative and integrable function on $[0,+\infty)$ that will be referred to as the $d$-radial Schoenberg density of $C$. 
On the other hand, any member $C$ of $\Phi_d$ such that $C(\| \cdot \|_d)$ is absolutely integrable in $\R^d$ admits the following Fourier-Hankel representation \citep{Chiles2012}:
\begin{equation}
    \label{fourier1}
    C(h) = {(2\pi)^{\frac{d}{2}}} h^{1-\frac{d}{2}} \int_0^{+\infty} u^{\frac{d}{2}} J_{\frac{d}{2}-1}(u h) f_d(u) {\rm d}u, \qquad h > 0,
\end{equation}
with
\begin{equation*}
    %\label{fourier2}
    f_d(u) = \frac{1}{(2\pi)^{\frac{d}{2}}} u^{1-\frac{d}{2}} \int_0^{+\infty} h^{\frac{d}{2}} J_{\frac{d}{2}-1}(u h) C(h) {\rm d}h, \qquad u > 0,
\end{equation*}
where $f_d: (0,+\infty) \to [0,+\infty)$ is a mapping that will be referred to as the $d$-radial spectral density of $C$. Comparing (\ref{schoenberg2}) and (\ref{fourier1}) leads to the following identity:
\begin{equation}
\label{schoenberg2fourier}
g_d(u) = \frac{2\pi^{\frac{d}{2}}}{ \Gamma(\frac{d}{2})} u^{d-1} f_d(u), \quad u > 0.
\end{equation}

\subsection{Sobolev Spaces}

We consider the classical Sobolev space $H^{s}(\R^d)$ defined through
\begin{equation*} 
H^s(\R^d) = \Big \{ f \in L_2(\R^d)\; : \; \hat{f}(\cdot) \left ( 1+\| \cdot \|_d \right )^{s/2} \in L_2(\R^d) \Big \},
\end{equation*}
equipped with the inner product
\begin{equation} 
\label{inner}
\langle f,g \rangle_{H^s(\R^d)} = \frac{1}{(2 \pi)^{d/2}} \int_{\R^d} \frac{\hat{f}(\cdot) \hat{g}(\cdot) }{\left ( 1 + \|\cdot \|_d^2 \right )^{-s} } {\rm d } \cdot.
\end{equation}
One can recognize that this is identical to the inner product (\ref{eqHK}) under the special case $\widehat{K}(\boldsymbol{\omega})= \left ( 1 + \|\boldsymbol{\omega} \|_d\right )^{-s} $. When $s= \nu +d/2$ with $\nu>0$, this inner product corresponds precisely to the Mat\'ern kernel $K(h) = {\cal M}_{1,\nu,d}(h)$ \citep{porcu2024matern}, with
\begin{equation}
\label{matern}
{\cal M}_{a,\nu,d}(h) = \frac{2^{1-\nu}}{\Gamma(\nu)} \left(\frac{h}{a}\right)^{\nu} {K}_{\nu}\left(\frac{h}{a}\right), \qquad h \ge 0, \quad a>0, \quad \nu>0.
\end{equation}

Although the expression of the kernel (\ref{matern}) does not depend on $d$, we use it in the indices of parameters to emphasize the dimension of the Euclidean space under consideration. The kernel (\ref{matern}) actually belongs to $\Phi_{\infty}$.

By the Sobolev embedding theorem, the function space $H^{s}(\R^d)$ is contained in the space of continuous functions in $\R^d$. Arguments in \cite{Wen} in concert with a straight comparison between (\ref{eqHK}) and (\ref{inner}) provide the desired connection: if a kernel $K$ belonging to $\Phi_d$ has a spectral density $\widehat{K}$ such that there exists constants $0<c_1<c_2<+\infty$ with 
\begin{equation}
\label{bounded} 
 c_1 (1+ \|\cdot\|_d^2)^{-s} \le \widehat{K}(\cdot) \le c_2 (1+ \|\cdot\|_d^2)^{-s}, \qquad s > d/2, 
\end{equation}
then the reproducing kernel associated with $K$ is norm equivalent to the Sobolev space $H^s(\R^d)$. This is one of the reasons why the Mat\'ern kernel has been so popular in statistics, machine learning, and numerical analysis.

For a Gaussian random field in $\R^d$, we consider mean square differentiability in the classical sense  \citep{Adler:1981} and we adopt the traditional definition of fractal dimension as Hausdorff dimension \citep{falconer2014fractal}. Both of these properties are, for the case of Gaussian random fields, in one-to-one correspondence with the local properties of the associated covariance kernel. 

\section{The Class ${\cal H}$ of Generalized Hypergeometric Kernels}
\label{sec: construction}

\subsection{Establishing the New Class}

The following details the parametric family of kernels that motivates this paper. 

\noindent {\bf Theorem 1} (Generalized hypergeometric kernel)
%\label{newclass}
{\it Let $a, \alpha, \beta, \gamma \in \mathbb{R}_{>0}$, $d \in \mathbb{N}_{\geq 1}$ and $k \in \mathbb{N}$. Let $\btheta = (a,\alpha,\beta,\gamma,d,k)^{\top}$, with $\top$ denoting the transpose. 
The mapping ${\cal H}_{\btheta}: [0,+\infty) \to \R$, defined by
\begin{equation}
    \label{GeneralizedHypergeometric}
    {\cal H}_{\btheta}(h) =  \varpi
    \left ( \frac{h}{a} \right )^{2\alpha-d-2k} \textstyle\pFq{3}{2}{\alpha,1+\alpha-\beta,1+\alpha-\gamma}{1+\alpha-\frac{d}{2}-k,\alpha-k}{{\frac{h^2}{a^2}}} 
    + \; \textstyle\pFq{3}{2}{\frac{d}{2}+k,1+\frac{d}{2}+k-\beta,1+\frac{d}{2}+k-\gamma}{1+\frac{d}{2}+k-\alpha,\frac{d}{2}}{\frac{h^2}{a^2}},\hfill 
\end{equation} for $0 \leq h < a$, and $0$ otherwise, 
with $$ \varpi \equiv \varpi(\alpha,\beta,\gamma,d,k) = \frac{\Gamma(\alpha) \Gamma(\beta-\frac{d}{2}-k) \Gamma(\gamma-\frac{d}{2}-k) \Gamma(\frac{d}{2})\Gamma(\frac{d}{2}+k-\alpha)}{\Gamma(\frac{d}{2}+k) \Gamma(\alpha-\frac{d}{2}-k)\Gamma(\beta-\alpha)\Gamma(\gamma-\alpha) \Gamma(\alpha-k)},$$ belongs to the class $\Phi_d$ provided the following parametric restrictions are adopted: 
\begin{itemize}
    \item[(A.1)] $\alpha > \frac{d}{2}+k$;
    \item[(A.2)] $2(\beta-\alpha) (\gamma-\alpha) \geq \alpha$;
    \item[(A.3)] $2(\beta+\gamma) \geq 6 \alpha +1$;    
    \item[(A.4)] $\alpha-\frac{d}{2}-k \notin \mathbb{N}$.
\end{itemize}}
\hfill\BlackBox

Apart from the dimension $d$ of the space $\R^d$ where ${\cal H}_{\btheta}(\| \cdot \|_d)$ is positive definite, the class ${\cal H}_{\btheta}$ has $5$ parameters, and their role will be progressively illustrated below. Note that conditions (A.2) and (A.3) imply that both $\beta$ and $\gamma$ are greater than $\alpha$. \\

\noindent {\bf Theorem 2}
%\label{spectrum}
{\it Let ${\cal H}_{\btheta}$ be the kernel defined through (\ref{GeneralizedHypergeometric}) and let conditions (A.1) to (A.4) in Theorem 1 hold. Then, the function ${\cal H}_{\btheta}(\|\cdot\|_d)$ is absolutely integrable in $\R^d$ and possesses a uniquely determined $d$-radial spectral density, denoted $\widehat{{\cal H}}_{\btheta}$, admitting expression
\begin{equation}
\label{hygeodensity}
\widehat{{\cal H}}_{\btheta}(u) = \widehat{\varpi} \,      
a^{d+2k} \, u^{2k} \textstyle\pFq{1}{2}{\alpha}{\beta,\gamma}{-\frac{a^2 u^2}{4}}, \quad 
\end{equation} for $u \in [0,+\infty)$, }
with $$ \widehat{\varpi}\equiv \widehat{\varpi}(\alpha,\beta,\gamma,d,k)=\frac{\Gamma(\frac{d}{2})\Gamma(\alpha) \Gamma(\beta-\frac{d}{2}-k) \Gamma(\gamma-\frac{d}{2}-k)}{\pi^{\frac{d}{2}}2^{d+2k}\Gamma(\frac{d}{2}+k) \Gamma(\alpha-\frac{d}{2}-k) \Gamma(\beta) \Gamma(\gamma)}.$$
\hfill\BlackBox\\

\noindent {\bf Theorem 3}
%\label{sobolev}
{\it Let ${\cal H}_{\btheta}(\|\cdot\|_d)$ be the kernel defined through (\ref{GeneralizedHypergeometric}) and let conditions (A.1) to (A.4) in Theorem 1 hold. {If, additionally, the inequality in (A.3) is strict, then} ${\cal H}_{\btheta}(\|\cdot\|_d)$ is a reproducing kernel with RKHS that is norm equivalent to the Sobolev space $H^{\alpha-k}(\R^d)$.}
\hfill\BlackBox\\

This result, in concert with the findings proved in Appendix \ref{appProofs}, parameterizes the Sobolev spaces associated with most of the parametric classes of continuous correlation functions that have been proposed in earlier literature. Table \ref{tab:hygeo} is a resum{\'e} of these kernels, being all of them members of $\Phi_d$ (with, possibly, restrictions on $d$ as indicated in the table) and special cases or asymptotic cases of the class ${\cal H}_{\btheta}$. To understand the table, we split the cases into two classes: when $k=0$, the class ${\cal H}_{\btheta}$ reduces to the Gauss Hypergeometric kernel introduced by \cite{emery2022gauss} (see Proposition 1 next). Hence, we report all the special cases according to either $k=0$ or $k\ne 0$. In turn, for every class, we report (third column) the parametric restriction on ${\cal H}_{\btheta}$ that allows to attain the corresponding kernel as a special or asymptotic case. The fourth column allows to understand whether the specific result is being shown in this paper, or belongs to earlier literature. To provide further insight on this table, a graphical representation of the same in the form of diagramatic relation is reported in Figure \ref{figure_relations}. 

\begin{landscape}
\begin{table}[ht!] 
    \tiny
    \caption{Special cases in $\Phi_d$ from the class ${\cal H}$.}
    \label{tab:hygeo}
    \begin{tabular}{l l l l}
         \toprule
         Model & Submodel & Restrictions & Reference \\
         \midrule
         & Euclid's hat & $\alpha = \frac{d+1}{2}$, $\beta = \alpha+\frac{1}{2}$, $\gamma=2\alpha$ & \cite{matheron1965, schaback1995creating} \\
         & Triangular & $\alpha = 1$, $\beta = \frac{3}{2}$, $\gamma=2$, $d=1$ & \cite{Matern1960}\\
         & Circular & $\alpha = \frac{3}{2}$, $\beta = 2$, $\gamma=3$, $d = 2$ & \cite{Zubrzycki}, \cite{Matern1960}\\
         & Spherical & $\alpha = 2$, $\beta = \frac{5}{2}$, $\gamma=4$, $d = 3$ & \cite{Matern1960}\\
         & Pentaspherical & $\alpha = 3$, $\beta = \frac{7}{2}$, $\gamma=6$, $d = 5$ & \cite{Matern1960}\\
         \cmidrule(lr){2-4}
         & Upgraded Euclid's hat & $\alpha > \frac{d}{2}$, $\beta = \alpha+\frac{1}{2}$, $\gamma=2\alpha$ & \cite{matheron1965, wu} \\
         & Cubic & $\alpha = 3$, $\beta = \frac{7}{2}$, $\gamma=6$, $d = 3$ & \cite{Chiles1977}\\
         & Penta & $\alpha = 4$, $\beta = \frac{9}{2}$, $\gamma=8$, $d = 3$ & \cite{Chiles2012}\\
         \cmidrule(lr){2-4}
         ${\cal H}$ class & Generalized Wendland & $\beta-\alpha \geq \frac{\alpha}{2} > \frac{d}{4}$, $\gamma=\beta+\frac{1}{2}$ & \cite{gneiting2002compactly, zastavnyi2006} \\
          & Ordinary Wendland & $\alpha-\frac{d+1}{2} \in \mathbb{N}$, $\beta-\alpha \geq \frac{\alpha}{2}$, $\gamma=\beta+\frac{1}{2}$ & \cite{gneiting1999atmospheric}\\
          & Original Wendland & $\alpha-\frac{d+1}{2} \in \mathbb{N}$, $2(\beta-\alpha) \in \mathbb{N}_{\geq \alpha}$, $\gamma=\beta+\frac{1}{2}$ & \cite{Wen}\\
         ($k=0$) & Missing Wendland & $\alpha-\frac{d}{2} \in \mathbb{N}_{\geq 1}$, $2(\beta-\alpha) \in \mathbb{N}_{\geq \alpha}$, $\gamma=\beta+\frac{1}{2}$ & \cite{Schaback2011}\\
         (Proposition 1) & Askey & $\alpha=\frac{d+1}{2}$, $\beta-\alpha \geq \frac{\alpha}{2}$, $\gamma=\beta+\frac{1}{2}$ & \cite{Golubov}\\
         & Quadratic & $\alpha=2, \beta=3$, $\gamma=\frac{7}{2}$, $d = 3$ & \cite{Alfaro1984}\\
         \cmidrule(lr){2-4}
         & Truncated power & $\alpha - \frac{d}{2} \in \mathbb{R}_{>0} \smallsetminus \mathbb{N}$, $\beta-\alpha \in \mathbb{N}_{\geq 1}$, $\gamma-\frac{d}{2} \in \mathbb{N}_{\geq 1}$ & \cite{emery2022gauss} \\
         & Truncated polynomial & $\alpha - \frac{d+1}{2} \in \mathbb{N}$, $\beta-\alpha \in \mathbb{N}_{\geq 1}$, $\gamma-\frac{d}{2} \in \mathbb{N}_{\geq 1}$  & \cite{emery2022gauss}\\
         \cmidrule(lr){2-4}
         & Mat\'ern & $\alpha> \frac{d}{2}$, $\beta \to +\infty$, $\gamma=\beta+\frac{1}{2}$, $a = 2\beta b$  & Proposition 7, \cite{Matern1960}\\
         & Exponential & $\alpha = \frac{d+1}{2}$, $\beta \to +\infty$, $\gamma=\beta+\frac{1}{2}$, $a = 2\beta b$  &  Proposition 7, \cite{Matern1960}\\
         \cmidrule(lr){2-4}
         & Gaussian & $\alpha \to +\infty$, $\beta/\alpha \to +\infty$, $\gamma=\beta+\frac{1}{2}$, $a = \beta b/\sqrt{\alpha}$  & Proposition 10, \cite{Matern1960}\\
         \cmidrule(lr){2-4}
         & Incomplete gamma & $\alpha > \frac{d}{2}$, $\beta>\alpha$, $\gamma \to + \infty$, $a = b \sqrt{\gamma}$  & \cite{emery2022gauss}\\
         & Gaussian-polynomial & $\alpha-\frac{d}{2}-1 \in \mathbb{N}$, $\beta>\alpha$, $\gamma \to + \infty$, $a = b \sqrt{\gamma}$  & Proposition 14\\
         & Complementary error & $\alpha = \frac{d+1}{2}$, $\beta > \alpha$, $\gamma \to + \infty$, $a = b \sqrt{\gamma}$  & \cite{dalenius}, \cite{gneiting1999radial}\\
         \midrule
         & Hole effect truncated power & $\alpha - \frac{d}{2} - k \in \mathbb{R}_{>0} \smallsetminus \mathbb{N}$, $\beta-\alpha \in \mathbb{N}_{\geq 1}$, $\gamma-\frac{d}{2}-k \in \mathbb{N}_{\geq 1}$ & Proposition 2\\
         & Hole effect truncated polynomial & $\alpha - \frac{d+1}{2}-k \in \mathbb{N}$, $\beta-\alpha \in \mathbb{N}_{\geq 1}$, $\gamma-\frac{d}{2}-k \in \mathbb{N}_{\geq 1}$ & Proposition 2\\
         \cmidrule(lr){2-4}
         ${\cal H}$ class & Hole effect Generalized Wendland & $\alpha > \frac{d}{2}+k$, $\beta-\alpha \geq \frac{\alpha}{2}$, $\gamma=\beta+\frac{1}{2}$ & Proposition 4, \cite{emery2025} \\
          & Hole effect ordinary Wendland & $\alpha - \frac{d+1}{2}-k \in \mathbb{N}$, $\beta-\alpha \geq \frac{\alpha}{2}$, $\gamma=\beta+\frac{1}{2}$ & Proposition 4, \cite{emery2025}  \\
          & Hole effect original Wendland & $\alpha - \frac{d+1}{2}-k \in \mathbb{N}$, $2(\beta-\alpha) \in \mathbb{N}_{\geq \alpha}$, $\gamma=\beta+\frac{1}{2}$ & Proposition 4, \cite{emery2025}  \\
         ($k \in \mathbb{N}_{\geq 1}$) & Hole effect Askey & $\alpha = \frac{d+1}{2}+k $, $\beta-\alpha \geq \frac{\alpha}{2}$, $\gamma=\beta+\frac{1}{2}$ & Proposition 6, \cite{emery2025} \\
         \cmidrule(lr){2-4}
         (Theorem 1) & Hole effect Mat\'ern & $\alpha> \frac{d}{2}+k$, $\beta \to +\infty$, $\gamma=\beta+\frac{1}{2}$, $a = 2\beta b$  & Proposition 8, \cite{emery2025}\\
         \cmidrule(lr){2-4}
         & Hole effect Gaussian & $\alpha \to +\infty$, $\beta/\alpha \to +\infty$, $\gamma=\beta+\frac{1}{2}$, $a = \beta b/\sqrt{\alpha}$ & Proposition 12\\
         \cmidrule(lr){2-4}
         & Schoenberg & $\alpha = \frac{d+1}{2}+2k$, $k \to +\infty$, $\frac{\beta}{k} \to +\infty$, $\gamma=\beta+\frac{1}{2}$, $a = 2\beta b$  & Proposition 9, \cite{schoenberg1938metric2} \\
         & Cosine &  $\alpha = 2k+1$, $k \to +\infty$, $\frac{\beta}{k} \to +\infty$, $\gamma=\beta+\frac{1}{2}$, $a = 2\beta b$, $d = 1$ & Proposition 9, \cite{Yaglom:1987}\\
         & Cardinal sine &  $\alpha = 2k+2$, $k \to +\infty$, $\frac{\beta}{k} \to +\infty$, $\gamma=\beta+\frac{1}{2}$, $a = 2\beta b$, $d=3$ & Proposition 9, \cite{Yaglom:1987}\\
         \cmidrule(lr){2-4}
         & Hole effect incomplete gamma & $\alpha > \frac{d}{2}+k$, $\beta>\alpha$, $\gamma \to + \infty$, $a = b \sqrt{\gamma}$  & Proposition 14\\
         \midrule
    \end{tabular}
\end{table}

\begin{figure}[H]
\begin{tikzpicture}[node distance={22.5mm}, thick, main/.style = {draw, rectangle}] 
\node[main] (1) [fill=green!70,text width=2.7cm] {Generalized hypergeometric};
\node[main] (2) [below of=1,fill=blue!15,text width=2.5cm] {Gauss \\hypergeometric};
\node[main] (3) [fill=blue!15,below right of=2,text width=2.0cm] {Generalized Wendland};
\node[main] (4) [fill=blue!15,below left of=2,text width=1.7cm] {Truncated power};
\node[main] (5) [fill=green!15,above right of=3,text width=2cm] {Hole effect Generalized Wendland};
\node[main] (6) [fill=blue!15,below of=3,text width=1.7cm] {Missing Wendland};
\node[main] (7) [fill=blue!15,below of=4,text width=1.9cm] {Truncated polynomial};
\node[main] (8) [fill=blue!15,right of=6,text width=1.7cm] {Ordinary Wendland};
\node[main] (9) [fill=blue!15,below left of=8,text width=1.7cm] {Original Wendland};
\node[main] (10) [fill=blue!15,right of=9] {Askey};
\node[main] (11) [fill=blue!15,below left of=9] {Quadratic};
\node[main] (12) [fill=green!15,above right of=8,text width=1.85cm] {Hole effect ordinary Wendland};
\node[main] (13) [right of=12] {Mat\'ern};
\node[main] (14) [fill=green!15,below right of=12,text width=1.9cm] {Hole effect Askey};
\node[main] (15) [fill=blue!15,left of=4] {Penta};
\node[main] (16) [fill=blue!15,above left of=15,text width=1.7cm] {Upgraded Euclid's hat};
\node[main] (17) [fill=blue!15,left of=15] {Cubic};
\node[main] (18) [fill=blue!15,below right of=17] {Triangular};
\node[main] (19) [fill=blue!15,left of=18,text width=1.3cm] {Euclid's hat};
\node[main] (20) [fill=blue!15,below left of=7] {Spherical};
\node[main] (21) [fill=blue!15,below of=19] {Circular};
\node[main] (22) [fill=blue!15,below left of=20] {Pentaspherical};
\node[main] (23) [fill=green!15,above right of=16,text width=1.9cm] {Hole effect truncated polynomial};
\node[main] (24) [fill=yellow!15,above right of=12,text width=1.8cm] {Hole effect Mat\'ern};
\node[main] (25) [fill=yellow!15,above right of=24] {Cosine};
\node[main] (26) [fill=yellow!15,below right of=25,text width=1.8cm] {Hole effect Gaussian};
\node[main] (27) [below right of=26] {Gaussian};
\node[main] (28) [below left of=27] {Exponential};
\node[main] (29) [fill=yellow!15,above of=26] {Schoenberg};
\node[main] (30) [fill=yellow!15,left of=25,text width=1.4cm] {Cardinal sine};
\node[main] (31) [fill=yellow!15,above of=5,text width=1.9cm] {Hole effect incomplete gamma};
\node[main] (32) [above left of=25,text width=1.8cm] {Incomplete gamma};
\node[main] (33) [above left of=23,text width=2.6cm] {Complementary error (erfc)};
\node[main] (34) [above right of=29,text width=1.8cm] {Gaussian-polynomial};
\draw[->] (1) -- (2);
\draw[->] (1) -- (5);
\draw[draw=blue,->] (2) -- (16);
\draw[draw=blue,->] (2) -- (4);
\draw[draw=blue,->] (2) -- (3);
\draw[draw=blue,->] (16) -- (15);
\draw[draw=blue,->] (16) -- (17);
\draw[draw=blue,->] (16) to [out=180,in=100,looseness=1] (19);
\draw[draw=blue,->] (19) -- (18);
\draw[draw=blue,->] (19) -- (20);
\draw[draw=blue,->] (19) -- (22);
\draw[draw=blue,->] (19) -- (21);
\draw[draw=blue,->] (4) -- (7);
\draw[draw=blue,->] (3) -- (6);
\draw[draw=blue,->] (12) -- (8);
\draw[draw=blue,->] (3) -- (8);
\draw[draw=blue,->] (8) -- (9);
\draw[draw=blue,->] (8) -- (10);
\draw[draw=blue,->] (9) -- (11);
\draw[draw=blue,->] (10) to [out=250,in=0,looseness=1.0] (11);
\draw[draw=blue,->] (7) -- (9);
\draw[draw=blue,->] (7) -- (15);
\draw[draw=blue,->] (7) -- (17);
\draw[draw=blue,->] (7) -- (18);
\draw[draw=blue,->] (7) -- (20);
\draw[draw=blue,->] (7) to [out=270,in=0,looseness=1.0] (22);
\draw[draw=blue,->] (7) -- (11);
\draw[draw=blue,->] (12) -- (14);
\draw[->] (1) -- (23);
\draw[draw=blue,->] (5) -- (3);
\draw[draw=blue,->] (5) -- (12);
\draw[draw=blue,->] (13) -- (28);
\draw[draw=blue,->] (29) -- (25);
\draw[draw=blue,->] (29) to [out=180,in=15,looseness=0.5] (30);
\draw[->] (26) -- (27);
\draw[->] (23) to [out=280,in=125,looseness=1.1] (7);
\draw[dashed][->] (26) -- (29);
\draw[dashed][draw=blue,->] (13) -- (27);
\draw[draw=blue,->] (14) -- (10);
\draw[dashed][draw=blue,->] (3) to [out=10,in=150,looseness=1.2] (13);
\draw[draw=blue,dashed][->] (5) -- (24);
\draw[dashed][draw=blue,->] (10) to [out=0,in=220,looseness=1.0] (28);
\draw[dashed][draw=blue,->] (29) to [out=320,in=90,looseness=1] (27);
\draw[draw=blue,->] (24) -- (13);
\draw[dashed][->] (24) -- (26);
\draw[dashed][->] (24) to [out=15,in=230,looseness=1] (29);
\draw[->] (32) -- (34);
\draw[->] (34) to [out=290,in=70,looseness=0.4] (27);
\draw[draw=blue,->] (32) to [out=145,in=0,looseness=0.1] (33);
\draw[->] (31) -- (32);
\draw[dashed][->] (1) -- (31);
\draw[dashed][draw=blue,->] (19) to [out=120,in=230,looseness=0.8] (33);

\end{tikzpicture} 
%}
\caption{Connections between covariance kernels. Blue boxes are compactly-supported kernels; yellow boxes are hole effect kernels; green boxes are compactly-supported hole effect kernels. %; gray boxes are generalized covariance kernels. 
Solid arrows indicate particular cases; dashed arrows indicate asymptotic cases. Connections established in previous literature are indicated in blue; connections proved in this paper are indicated in black.}
\label{figure_relations}
\end{figure}
\end{landscape} 

\subsection{Properties} 
\label{local}

\noindent \textbf{Support.} ${\cal H}_{\btheta}$ is compactly supported, as it vanishes outside the interval $[0,a)$.\\

\noindent \textbf{Hole effect.} ${\cal H}_{\btheta}$ is nonnegative and monotonic when $k=0$ \citep{emery2022gauss}, but attains negative values when $k>0$, as shown in Appendix \ref{walk} and illustrated hereinafter.\\

\noindent \textbf{Smoothness.}
By using formula 16.3.1 in \cite{Olver}, one finds the first- and second-order right derivatives of ${\cal H}_{\btheta}$ at $h=0$:
\begin{equation*}
    \begin{split}
    &\frac{\partial_+ {\cal H}_{\btheta}(h)}{\partial h}\Bigg \vert_{h=0} = \begin{cases}
        0 \text{ if $2\alpha > d+2k+1$}\\
        -\frac{2\Gamma(\alpha) \Gamma(\beta-\alpha+\frac{1}{2}) \Gamma(\gamma-\alpha+\frac{1}{2}) \Gamma(\frac{d}{2})}{a \Gamma(\alpha-\frac{1}{2}) \Gamma(\beta-\alpha)\Gamma(\gamma-\alpha) \Gamma(\frac{d+1}{2})} \text{ if $2\alpha = d+2k+1$}\\
        -\infty \text{ if $2\alpha < d+2k+1$}.
    \end{cases}\\    
    &\frac{\partial^2_+ {\cal H}_{\btheta}(h)}{\partial h^2} \Bigg \vert_{h=0} = \begin{cases}
        \frac{4(\frac{d}{2}+k)(1+\frac{d}{2}+k-\beta)(1+\frac{d}{2}+k-\gamma)}{(1+\frac{d}{2}+k-\alpha) d a^2} \text{ if $2\alpha > d+2k+2$}\\
        +\infty \text{ if $2\alpha < d+2k+2$}.
    \end{cases}
    \end{split}    
\end{equation*}
Note that the case $2\alpha = d+2k+2$ is excluded by condition (A.4).

Accordingly, the parameter $\alpha-k$ controls the regularity of ${\cal H}_{\btheta}$ at the origin. When $\alpha-k > \frac{d}{2}+1$, $h \mapsto {\cal H}_{\btheta}(h)$ is twice differentiable at $h=0$ and is associated with a random field that is mean square differentiable in space. 

More generally, by expanding the generalized hypergeometric function ${}_3F_2$ in (\ref{GeneralizedHypergeometric}) into a power series, one obtains an expansion of ${ {\cal H}_{\btheta}(h)}$ whose most irregular term is $h^{2\alpha-d-2k}$, where the exponent $2\alpha-d-2k$ is not an even integer due to condition (A.4). The function ${ {\cal H}_{\btheta}}$ therefore admits finite right derivatives at $h=0$ up to order $\lfloor 2\alpha-d-2k \rfloor$, with the odd-order derivatives being zero up to order $\lceil 2\alpha-d-2k-1 \rceil$. This implies that ${\cal H}_{\btheta}$ is associated with a random field that is $\lfloor \alpha-\frac{d}{2}-k \rfloor$-times mean square differentiable in space.\\

\noindent \textbf{Behavior near the range.} 
${ {\cal H}_{\btheta}}$ is continuous on $[0,+\infty)$ and infinitely differentiable on $(0,a) \cup (a,+\infty)$. In particular, it is continuous at $h=a$, but it may not be differentiable at this particular point. A sufficient condition for ${\cal H}_{\btheta}$ to be $p$-times differentiable at $h=a$ is that $\beta+\gamma-\alpha-2k-\frac{d}{2}-1>p$; this condition is also necessary when $\beta + \gamma \notin \mathbb{N}$ and $\beta + \gamma - \alpha - \frac{d}{2}\notin \mathbb{N}$ (Appendix \ref{alternativesGH}).\\

%As an example, let $\btheta=(a,\alpha,\frac{3}{4}+\frac{\alpha}{2},\frac{5}{4}+\frac{\alpha}{2},1,0)^{\top}$ with $\alpha \in (\frac{1}{2},3-\frac{\sqrt{21}}{2}]$, which satisfies the validity conditions of Theorem 1. Accounting for (\ref{GeneralizedHypergeometric2}) (see Appendix \ref{alternativesGH}) and for \citet[7.3.1.107]{prud} and \citet[15.8.4]{Olver}, one obtains the following closed-form expression of ${\cal H}_{\btheta}$:
%\begin{equation*}
%%\label{newd1}
%{\cal H}_{\btheta}(h)= \begin{cases} 2^{\frac{1}{2}-\alpha} \left[\left(1+\sqrt{1-\frac{h^2}{a^2}} \right)^{\alpha-\frac{1}{2}} - \left(1-\sqrt{1-\frac{h^2}{a^2}} \right)^{\alpha-\frac{1}{2}} \right], & 0 \leq h < a\\ 0, & h \geq a. \end{cases}
%\end{equation*}
%An illustration is given in Figure \ref{fig:modeloraro}. Note the irregular behavior (one-sided differentiability) of the kernel at the origin and at the range. Also, as $\alpha$ tends to $\frac{1}{2}$, the kernel tends pointwise to a pure nugget correlation function.
%\begin{figure}[H]
    %\centering
%\includegraphics[width = 0.45\textwidth]{modeloraro.png}
    %\caption{${\cal H}_{\btheta}$ for $\btheta=(a,\alpha,\frac{3}{4}+\frac{\alpha}{2},\frac{5}{4}+\frac{\alpha}{2},1,0)^{\top}$ for different choices of $\alpha$.}
    %\label{fig:modeloraro}
%\end{figure}

\noindent \textbf{Continuation.} 
Formula (\ref{GeneralizedHypergeometric}) is undefined if $\alpha-\frac{d}{2}-k \in \mathbb{N}_{\geq 1}$, as it involves the difference of two infinite terms. However, in such a case, the generalized hypergeometric kernel can be defined by continuation (proof in Appendix \ref{alternativesGH}): 
\begin{equation}
\label{continuationHG}
    \begin{split}
    {\cal H}_{\btheta}(h) = \lim_{\varepsilon \to 0^-} {\cal H}_{\btheta + (0,\varepsilon,0,0,0,0)^{\top}}(h), \quad h \geq 0, \quad \alpha-\frac{d}{2}-k \in \mathbb{N}_{\geq 1}.
    \end{split}
\end{equation}
Such a limit kernel belongs to $\Phi_d$.\\

\noindent \textbf{Fractal dimension.} Owing to Tauberian theorems, a Gaussian random field with covariance kernel ${\cal H}_{\btheta}$ has realizations with fractal dimension $D=d+1-\frac{\vartheta}{2}$ whenever $0 < \vartheta=2(\alpha-k)-1 \le 2$. To prove it, one just needs to observe that $\widehat{{\cal H}}_{\btheta}(u)$ behaves like $u^{-\vartheta-1}$ as $u \to +\infty$ (see proof of Theorem 3).

\section{Special Cases through Exact Parameterization} \label{sec: special}

\noindent {\bf Proposition 1} (Gauss hypergeometric kernel)
%\label{hygeocov}
{\it Let $\btheta=(a,\alpha,\beta,\gamma,d,0)^{\top}$ satisfying conditions (A.1) to (A.4) as per Theorem 1. Then, ${\cal H}_{\btheta}$ is the Gauss hypergeometric kernel introduced by \cite{emery2022gauss}:
\begin{equation}
    \label{hygeo2F1}
    {\cal H}_{\btheta}(h)
    {\frac{\Gamma(\beta-\frac{d}{2})\Gamma(\gamma-\frac{d}{2}) }{\Gamma(\beta-\alpha+\gamma-\frac{d}{2}) \Gamma(\alpha-\frac{d}{2})}} \; \left(1-\frac{h^2}{a^2}\right)_+^{\beta-\alpha+\gamma-\frac{d}{2}-1} \, \textstyle\pFq{2}{1}{\beta-\alpha,\gamma-\alpha}{\beta-\alpha+\gamma-\frac{d}{2}}{1-\frac{h^2}{a^2}}.  
\end{equation}}
\hfill\BlackBox\\

The kernel (\ref{hygeo2F1}) is well defined and belongs to $\Phi_d$ even if condition (A.4) does not hold.\\

\noindent {\bf Proposition 2} (Hole effect truncated power and truncated polynomial kernels)
%\label{holeeffecttruncated}
{\it Let $\btheta=(a,\alpha,\beta,\gamma,d,k)^{\top}$ satisfying conditions (A.1) to (A.4) as per Theorem 1, such that $\beta = 1+\alpha+M$ and $\gamma=1+\frac{d}{2}+k+N$ with $M, N \in \mathbb{N}$. Then, one has
\begin{equation}
\label{truncatedpow}
   {\cal H}_{\btheta}(h) 
   = \begin{cases}
   0 \hfill \text{ if $a \leq h$,}\\
   \sum_{n=0}^N \frac{(\frac{d}{2}+k)_n (\frac{d}{2}+k-\alpha-M)_n (-N)_n}{(1+\frac{d}{2}+k-\alpha)_n (\frac{d}{2})_n n!} {\left ( \frac{h}{a} \right ) ^{2n}} \\ +\frac{\Gamma(\alpha) \Gamma(1+\alpha+M-\frac{d}{2}-k) \Gamma(\frac{d}{2})\Gamma(\frac{d}{2}+k-\alpha) N!}{\Gamma(\frac{d}{2}+k) \Gamma(\alpha-\frac{d}{2}-k)\Gamma(1+\frac{d}{2}+k+N-\alpha) \Gamma(\alpha-k) M!} \\    \times \sum_{n=0}^M \frac{(\alpha)_n (-M)_n (\alpha-\frac{d}{2}-k-N)_n}{(1+\alpha-\frac{d}{2}-k)_n (\alpha-k)_n n!} \left ( \frac{h}{a} \right )^{2n+2\alpha-d-2k} \hfill \text{ if $0 \leq h < a$.}\\
   \end{cases}
\end{equation}
If, additionally, $2\alpha - d$ is an odd integer, ${\cal H}_{\btheta}$ reduces to a truncated polynomial function. The related permissibility conditions (A.1) to (A.4) become
\begin{itemize}
    \item $\alpha = \frac{d+1}{2}+k+p$ with $p \in \mathbb{N}$;
   % \item $\beta = 1+\alpha+M$;
   % \item $\gamma = 1+\frac{d}{2}+k+N$;
    \item $(1+M) (2N-d-2k-4p) \geq \frac{d+1}{2}+k+p$;
    \item $M+N-k-2p \geq \frac{d-1}{2}$.    
\end{itemize}}
\hfill\BlackBox\\

\begin{figure}
    \centering
\includegraphics[width = 0.99\textwidth]{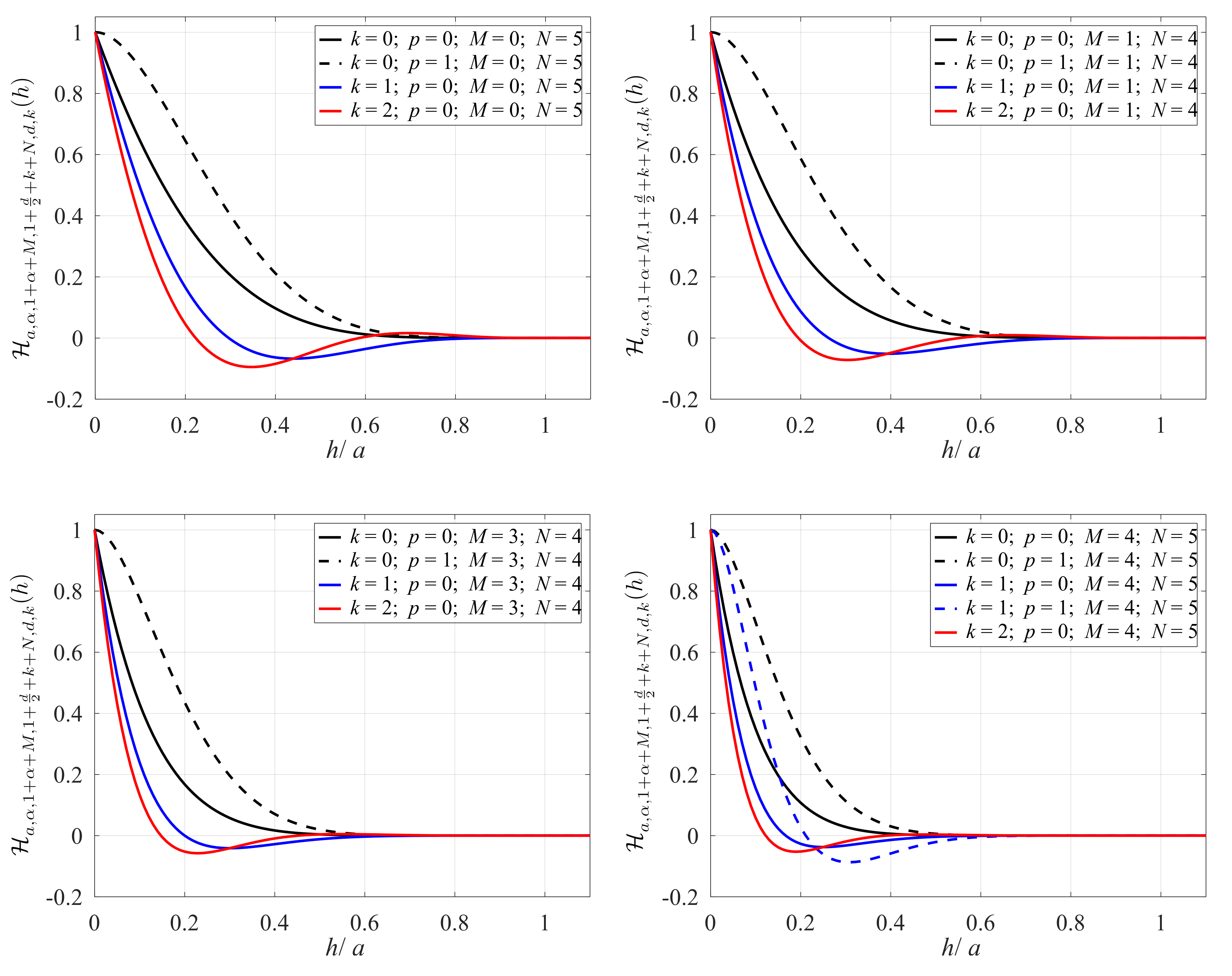}
    \caption{Hole effect truncated polynomial kernel ${\cal H}_{\btheta}$ for $\btheta=(a,\frac{d+1}{2}+k+p,1+\frac{d+1}{2}+k+p+M,1+\frac{d}{2}+k+N,d,k)^{\top}$, for different choices of $k$, $p$, $M$ and $N$.}
    \label{fig:fig1}
\end{figure}

Figure \ref{fig:fig1} gives examples of such a truncated polynomial kernel (\ref{truncatedpow}) with $2\alpha - d$ an odd integer in the two-dimensional space ($d=2$), for $p\leq 1$ and several choices of the other parameters $k$, $M$ and $N$. One observes that the behavior at the origin gets more regular as $p$ increases, which corresponds to a random field getting smoother in space. Also, a hole effect emerges with $k>0$, while the case $k=0$ (Gauss hypergeometric kernel) provides mappings that are monotonically nonincreasing.  \\

\noindent {\bf Proposition 3} (Ordinary and Generalized Wendland kernels)
%\label{noholeeffectGW}
{\it The Generalized Wendland kernel \citep{zastavnyi2006} is a special case of the Gauss hypergeometric kernel (\ref{hygeo2F1}), for which one has ${\cal W}_{a,\xi,\nu,d,0}={\cal H}_{\btheta}$ with $\btheta=(a,\xi+\frac{d+1}{2},\xi+\frac{d+\nu+1}{2}, \xi+\frac{d+\nu}{2}+1,0 )^{\top}$. In particular, 
\begin{equation}
\label{geneWendl}
{\cal W}_{a,\xi,\nu,d,0}(h) 
= \frac{\Gamma(\xi+\frac{\nu+1}{2})\Gamma(\xi+\frac{\nu}{2}+1) \left(1-\frac{h^2}{a^2}\right)_+^{\xi+\nu}}{\Gamma(\xi+\nu+1) \Gamma(\xi+\frac{1}{2})}  \, \textstyle\pFq{2}{1}{\frac{\nu}{2},\frac{\nu+1}{2}}{\xi+\nu+1}{\left(1-\frac{h^2}{a^2} \right)_+},
\end{equation}
with $\xi > -\frac{1}{2}$ and $\nu \geq \nu_{\min}(\xi,d)$, where
\begin{equation*}
    \nu_{\min}(\xi,d) := 
    \begin{cases}
    \frac{\sqrt{8\xi+9}-1}{2} \text{ if $d = 1$ and $-\frac{1}{2}<\xi<0$}\\
    \xi+ \frac{d+1}{2} \text{ otherwise.}
    \end{cases}
\end{equation*}}
\hfill\BlackBox 

The case when $\xi$ is an integer is known as the \emph{ordinary} Wendland kernel, for which a closed-form expression is available (see \citealp{hubbert2012closed} and \citealp{bevi2024}). The subcase when both $\xi$ and $\nu$ are integers yields the so-called \emph{original} Wendland kernel, which has a polynomial expression in the interval $[0,a]$ \citep{Wen}. The case when $\xi$ is a half-integer and $\nu$ is an integer is known as the \emph{missing} Wendland kernel \citep{Schaback2011}, which also has a closed-form expression \citep{bevi2024}. \\

\noindent {\bf Proposition 4} (Hole effect Generalized Wendland kernel)
%\label{holeeffectGW}
{\it The hole effect Generalized Wendland kernel ${\cal W}_{a,\xi,\nu,d,k}$ \citep{emery2025} is a particular case of the generalized hypergeometric kernel, for which one has 
\begin{equation}
\label{wendlandext}
{\cal W}_{a,\xi,\nu,d,k}(h) := {\cal H}_{a, \xi+\frac{d+1}{2}+k,\xi+\frac{d+\nu+1}{2}+k,\xi+\frac{d+\nu}{2}+k+1,d,k}(h),
\end{equation}
with $k \in \mathbb{N}$, $\xi > -\frac{1}{2}$ and $\nu \geq \nu_{\min}(\xi,d+2k)$.}
\hfill\BlackBox \\

Note that ${\cal W}_{a,\xi,\nu,d,k}$ reduces to the Generalized Wendland kernel (\ref{geneWendl}) if $k=0$. Also, if $\xi+\frac{1}{2} \in \mathbb{N}$, one has to consider the continuation (\ref{continuationHG}) of the generalized hypergeometric kernel in (\ref{wendlandext}). 
Closed-form expressions of ${\cal W}_{a,\xi,\nu,d,k}$ can be obtained when $\xi \in \mathbb{N}$, which yields a \emph{hole effect ordinary Wendland} kernel, see \cite{emery2025}.\\

\noindent {\bf Proposition 5} (Askey kernel)
%\label{exAskey0}
{\it The Askey kernel $h \mapsto \left(1-\frac{h}{a}\right)_+^{\nu}$ \citep{Golubov} is a particular case of the generalized hypergeometric kernel, corresponding to ${\cal W}_{a,0,\nu,d,0}$ with $\nu \geq \frac{d+1}{2}$.}
\hfill\BlackBox \\

\noindent {\bf Proposition 6} (Hole effect Askey kernel)
%\label{exAskey}
{\it The hole effect Askey kernel \citep{emery2025} is a particular case of the generalized hypergeometric kernel, corresponding to ${\cal W}_{a,0,\nu,d,k}$ with $\nu \geq \frac{d+1}{2}+k$.}
\hfill\BlackBox \\

\begin{figure}[H]
\begin{center}
\includegraphics[width = 0.99\textwidth]{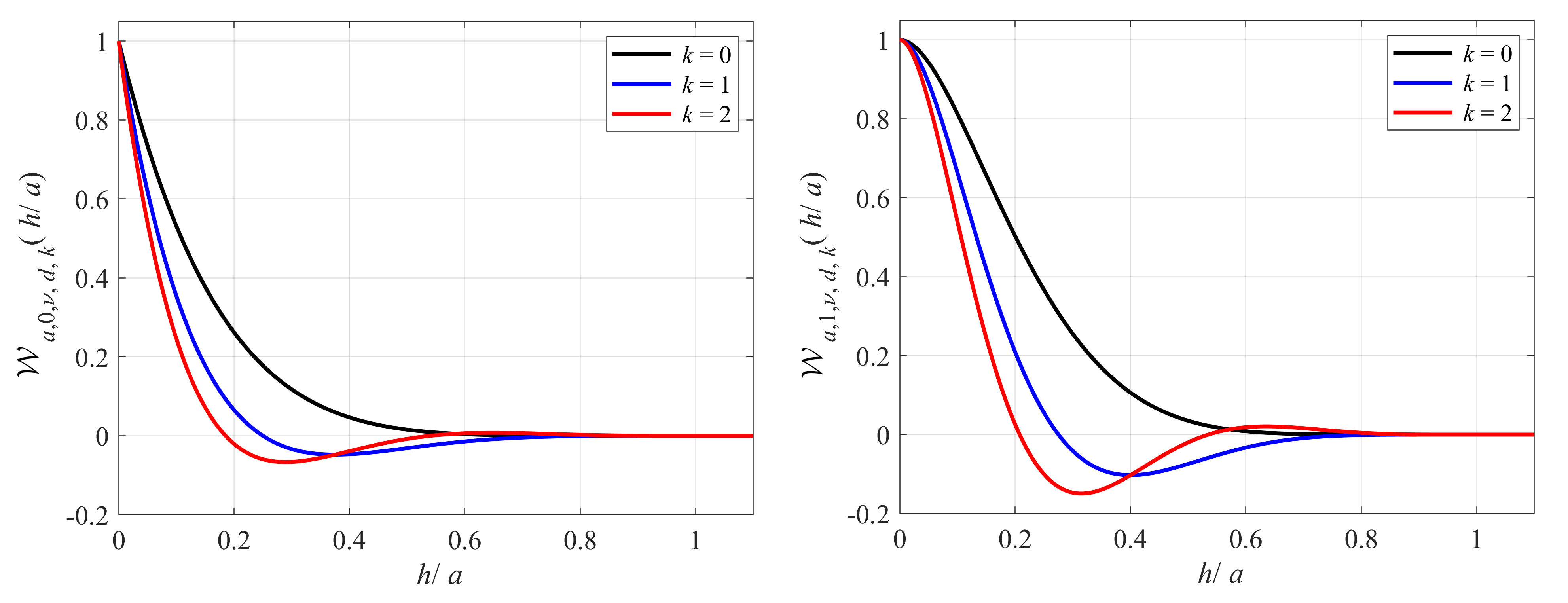}
\end{center}
\caption{Hole effect Askey ${\cal W}_{a,0,\nu,d,k}$ (left) and ordinary Wendland ${\cal W}_{a,1,\nu,d,k}$ (right) kernels, for $\nu = 6$, $d=2$ and $k = 0, 1, 2$.}
 \label{holegw}
\end{figure}

An illustration of the hole effect Askey and ordinary Wendland kernels is given in Figure \ref{holegw}. It can be appreciated that, as $k$ increases, the hole effect also increases. 

\section{Special Cases through Parametric Convergence} \label{sec: convergence}

\subsection{Mat\'ern-like Kernels}

The following result is of independent interest and provides a parameterization of the Generalized Wendland kernel that includes the Mat\'ern kernel as a limit case.\\

\noindent {\bf Proposition 7} 
%\label{wend2mat}
    {\it Let $a, \mu, \nu \in \mathbb{R}_{>0}$ and $d \in \mathbb{N}_{\geq 1}$. As $\mu$ tends to $+\infty$, the Generalized Wendland kernel ${\cal W}_{\mu a,\nu-\frac{1}{2},\mu,d,0}$ converges uniformly on $[0,+\infty)$ to the Mat\'ern kernel ${\cal M}_{a,\nu,d}$.}
\hfill\BlackBox \\

\noindent {\bf Proposition 8} 
\label{matgeneralized} 
{\it Let $a, \mu, \nu \in \mathbb{R}_{>0}$, $d \in \mathbb{N}_{\geq 1}$ and $k \in \mathbb{N}$. As $\mu$ tends to $+\infty$, the hole effect Generalized Wendland kernel ${\cal W}_{\mu a,\nu-\frac{1}{2},\mu,d,k}$ converges uniformly on $[0,+\infty)$ to the hole effect Mat\'ern kernel ${\cal M}_{a,\nu,d,k}$, defined through \citep{emery2025}
\begin{equation}
    \label{matgeneralized2}
\begin{split}
  {\cal M}_{a,\nu,d,k}(h) := \begin{cases}
  \sum_{q=0}^k \sum_{r=0}^{\max\{0,q-1\}} \sum_{s=0}^{q-r} \sum_{t=0}^{q-r-s} \left ( \frac{h}{a} \right )^{\nu+q-r-s} {K}_{\nu+2t+r+s-q}\left ( \frac{h}{a} \right )\\
  \quad \times \frac{(-1)^{q-s} (q-r)! (q-r)_r (\nu+1-s)_{s} (k-q+1)_q (q)_r}{2^{\nu+2q-s-1} q! \, r! \, s! \, t! \, (q-r-s-t)! \, \Gamma(\nu) (\frac{d}{2})_q} \text{ if $h>0$}\\
  1 \text{ if $h=0$.}
  \end{cases}
\end{split}
\end{equation}}
\hfill\BlackBox 

If $k=0$, then one recovers the traditional Mat\'ern kernel (\ref{matern}). Another special case is obtained when $\nu$ is a half-integer, in which case the modified Bessel functions can be expressed in terms of exponential and power functions \citep[8.468]{grad}. Analytical expressions of ${\cal M}_{a,\nu,d,k}$ in terms of modified Bessel functions of the first kind, generalized hypergeometric functions, or Meijer functions, can be found in \cite{emery2025}.

For $\nu \in \mathbb{N}_{\geq 1}$, Proposition 8 still holds by considering the continuation of the hole effect Generalized Wendland kernel \citep{emery2025}, while the hole effect Mat\'ern kernel remains given by (\ref{matgeneralized2}).

\begin{figure}[H]
\begin{center}
\includegraphics[width = 0.99\textwidth]{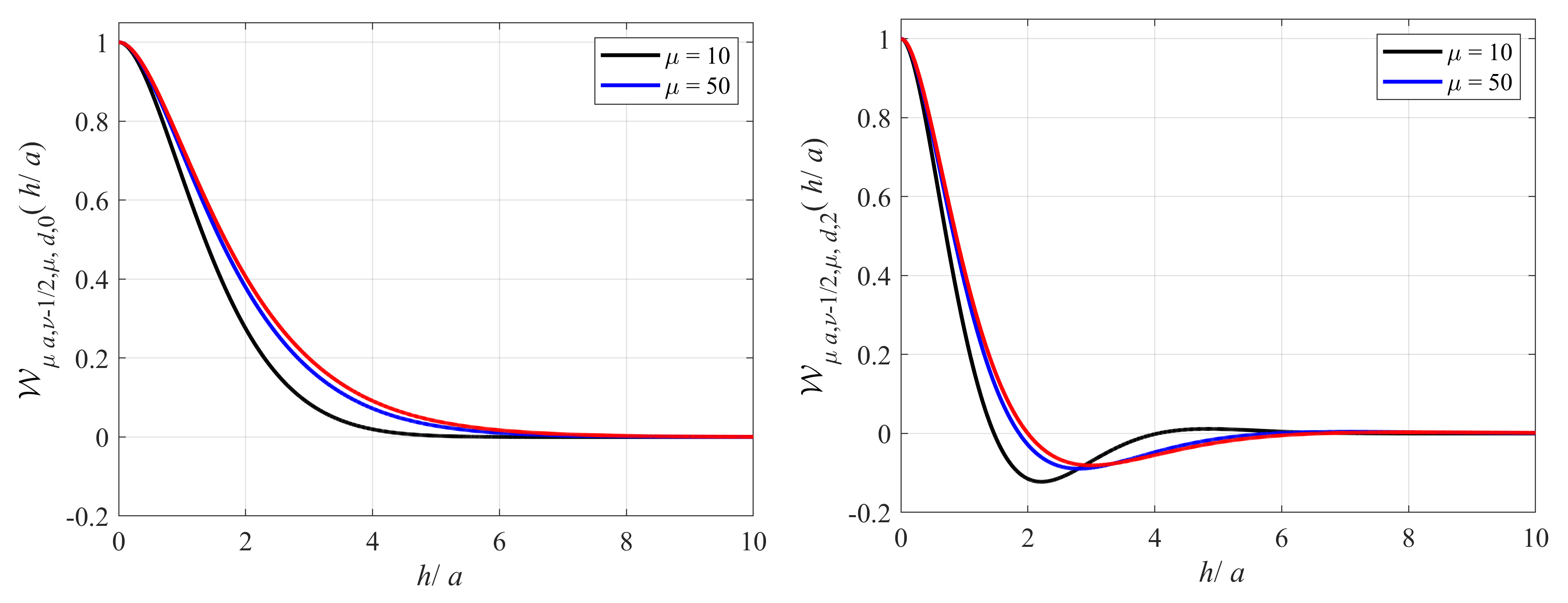}
\end{center}
\caption{${\cal W}_{\mu a,\nu-\frac{1}{2},\mu,d,k}$ and ${\cal M}_{a,\nu,d,k}$ (red line) when $\mu=10, 50$, $\nu = 1.5$, $d=2$ and $k=0$ (left) or $k=2$ (right).}
 \label{convergence}
\end{figure}

As an illustration, the convergence of ${\cal W}_{\mu a,\nu-\frac{1}{2},\mu,d,k}$ to ${\cal M}_{a,\nu,d,k}$ as $\mu$ tends to $+\infty$ can be appreciated in Figure \ref{convergence} when $k=0$ or $k=2$ for specific values of the parameters.

\subsection{Schoenberg Kernel}

When $k$ and $\nu$ increase at the same time, the kernels ${\cal W}_{\mu a,\nu-\frac{1}{2}, \mu, d,k}$ and ${\cal M}_{a,\nu,d,k}$ behave as differentiable (at the origin) oscillating correlation functions. In particular, the following result establishes the convergence of these kernels to the Schoenberg kernel (\ref{schoenbergJ}) as $k$ tends to infinity, which is an infinitely differentiable and oscillating correlation function that has infinitely many zeros \citep{Chiles2012}.\\

\noindent {\bf Proposition 9}
%\label{mat2schoen}
       {\it Let $a, \mu \in \mathbb{R}_{>0}$, $d \in \mathbb{N}_{\geq 1}$ and $k \in \mathbb{N}$. As both $k$ and $\frac{\mu}{k}$ tend to $+\infty$, the hole effect Generalized Wendland kernel ${\cal W}_{\mu a,k,\mu,d,k}$ and the hole effect Mat\'ern kernel ${\cal M}_{a,k + \frac{1}{2},d,k}$ converge uniformly on any bounded interval of $[0,+\infty)$ to the Schoenberg kernel ${\cal J}_{a,d}$.}
\hfill\BlackBox 

Particular cases of Schoenberg kernels include the cosine and cardinal sine kernels, for $d=1$ and $d=3$, respectively.

\begin{figure}[H]
\begin{center}
\includegraphics[width = 0.495\textwidth]{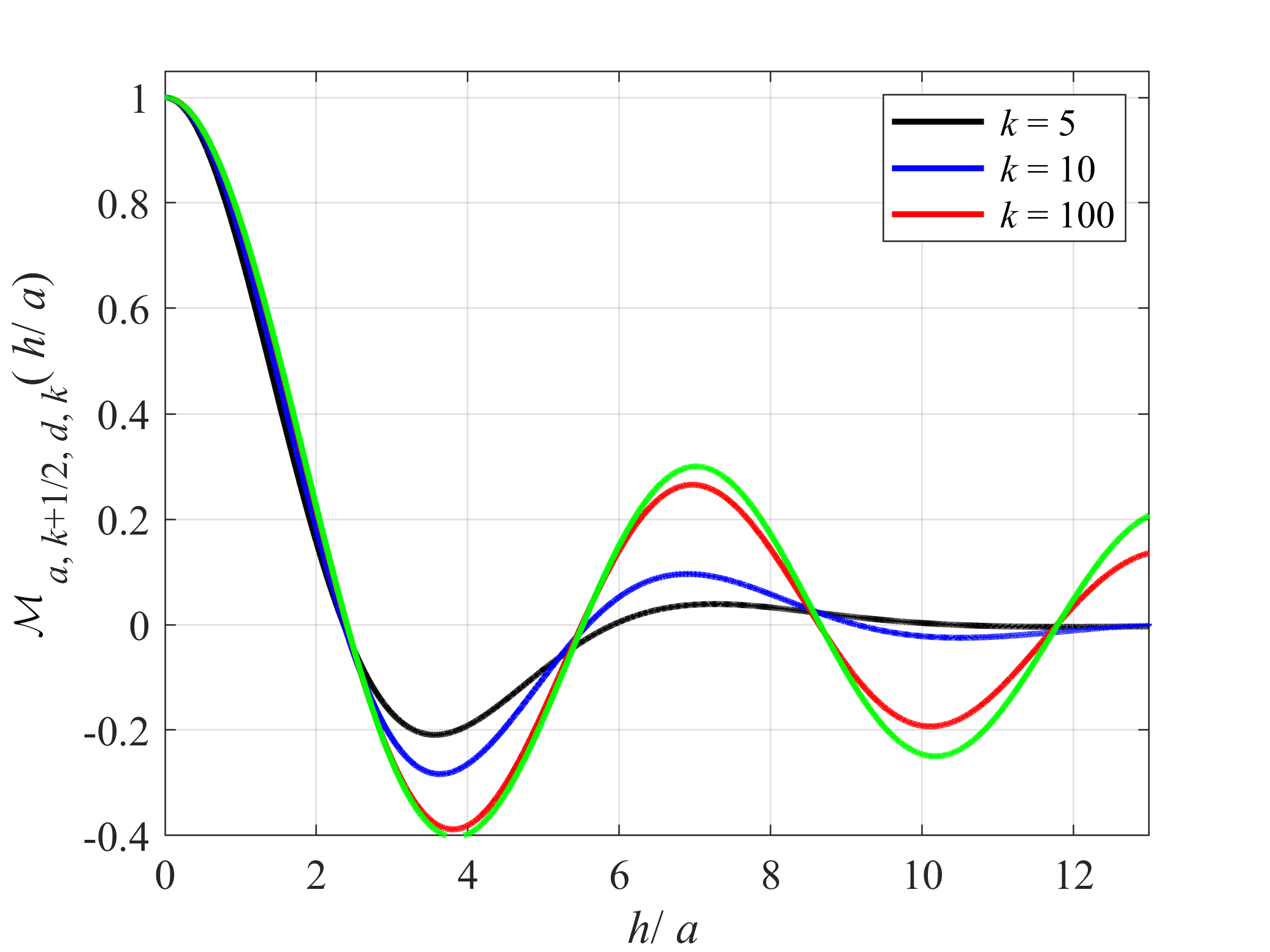}
\end{center}
\caption{${\cal M}_{a,k+\frac{1}{2},d,k}$ when $d=2$ and $k=5, 10, 100$ and ${\cal J}_{a,d}$ (green line).}
 \label{conve4}
\end{figure}

As an illustration, Figure \ref{conve4} depicts the ${\cal M}_{a,k + \frac{1}{2},d,k}$ kernel for $d=2$ and $k=5, 10, 100$ and the Schoenberg kernel ${\cal J}_{a,d}$. The former kernels tend to the latter as $k$ increases.

\subsection{Gaussian-like Kernels}

We also have convergence results to the well-known Gaussian kernel, which belongs to $\Phi_{\infty}$ and is defined as
\begin{equation}
    \label{gauss}
    {\cal G}_a (h) = \exp\left(-\frac{h^2}{a^2}\right), \quad h \geq 0, \quad a > 0.
\end{equation}

\noindent {\bf Proposition 10}
%\label{gaussian}
   {\it Let $a, \mu, \nu \in \mathbb{R}_{>0}$ and $d \in \mathbb{N}_{\geq 1}$. As both $\nu$ and $\mu \nu^{-2}$ tend to $+\infty$, the Generalized Wendland kernel ${\cal W}_{{\mu a}/\sqrt{4\nu},\nu-\frac{1}{2},\mu,d,0}$ uniformly converges on $[0,+\infty)$ to the Gaussian kernel ${\cal G}_a$.}
\hfill\BlackBox \\

Note that both the Mat\'ern and Schoenberg kernels also uniformly converge to the Gaussian kernel under a suitable parameterization, see \cite{stein-book} and \cite{schoenberg1938metric2}.\\

\noindent {\bf Proposition 11} (Hole effect Gaussian kernel)
%\label{wend2gaupol1}
{\it Let $a \in \mathbb{R}_{>0}$, $d \in \mathbb{N}_{\geq 1}$ and $k \in \mathbb{N}$. Define the hole effect Gaussian kernel as
\begin{equation}
\label{gaulaguerre}
    \begin{split}
    {\cal G}_{a,d,k}(h) := \frac{\Gamma(\frac{d}{2}) k!}{\Gamma(\frac{d}{2}+k)} \exp\left(-\frac{h^2}{4a^2}\right) L_k^{\frac{d}{2}-1}\left(\frac{h^2}{4a^2}\right), \quad h \geq 0,
    \end{split}
\end{equation}
with the generalized Laguerre polynomial $L_k^{\frac{d}{2}-1}$ given by \citep[5.5.3 and 18.5.12]{Olver}
\begin{equation}
\label{genlagpol}
    L_k^{\frac{d}{2}-1}(x) = \sum_{n=0}^{k} \frac{(\frac{d}{2}+n)_{k-n} (-x)^n}{(k-n)! n!} = \frac{\Gamma(k+\frac{d}{2})}{k!} \sum_{n=0}^{k} \frac{(-k)_n x^n}{\Gamma(n+\frac{d}{2}) n!}, \quad x \in \mathbb{R}.
\end{equation}
Then ${\cal G}_{a,d,k}$ belongs to $\Phi_d$.}
\hfill\BlackBox \\

The generalized Laguerre polynomial (\ref{genlagpol}) has $k$ different zeros that are positive \citep{Szego}, then so does ${\cal G}_{a,d,k}$: since it asymptotically tends to $0$ at infinity, this kernel has a hole effect with $k$ waves. Also note that (\ref{gauss}) is a particular case of (\ref{gaulaguerre}), as ${\cal G}_{a,d,0} = {\cal G}_{a}$.\\

\noindent {\bf Proposition 12} 
%\label{wend2gaupol2}
   {\it Let $a, \mu \in \mathbb{R}_{>0}$, $d, n \in \mathbb{N}_{\geq 1}$ and $k \in \mathbb{N}$. As both $n$ and $\frac{\mu}{n}$ tend to $+\infty$, ${\cal W}_{\mu a/\sqrt{4n},n,\mu,d,k}$ and ${\cal M}_{{a}/{\sqrt{4n}},n+\frac{1}{2},d,k}$ uniformly converge on any bounded interval of $[0,+\infty)$ to ${\cal G}_{a,d,k}$.}
\hfill\BlackBox \\

\noindent {\bf Proposition 13} 
%\label{gaupoltoSchoen}
    {\it Let $a \in \mathbb{R}_{>0}$, $d \in \mathbb{N}_{\geq 1}$ and $k \in \mathbb{N}$. As $k$ tends to $+\infty$, ${\cal G}_{a\sqrt{k},d,k}$ uniformly converges on any bounded interval of $[0,+\infty)$ to ${\cal J}_{a,d}$.}
\hfill\BlackBox \\

\subsection{Incomplete Gamma Kernels}

\noindent {\bf Proposition 14} (Hole effect incomplete gamma kernel)
%\label{incgamma}
{\it Let $\btheta = (a\sqrt{\gamma},\alpha,\beta,\gamma,d,k)^\top$ satisfying conditions (A.1) to (A.4) of Theorem 1. As $\gamma$ tends to $+\infty$, ${\cal H}_{\btheta}$ uniformly converges on any bounded interval of $[0,+\infty)$ to the hole effect incomplete gamma kernel ${\cal I}_{a,\alpha,d,k}$, defined by
\begin{equation*} 
\begin{split}    
    {\cal I}_{a,\alpha,d,k}(h) &= 1-\sum_{n=0}^k \frac{(-1)^n k! (\alpha-k+n)_{k-n}}{n! (k-n)! \Gamma(\alpha-\frac{d}{2}-k) \, (\frac{d}{2})_k} \Gamma^-\left(\alpha-\frac{d}{2}-k+n,\frac{h^2}{a^2}\right), \quad h \geq 0.
\end{split}
\end{equation*}}
\hfill\BlackBox \\
The name of the kernel is due to the fact that ${\cal I}_{a,\alpha,d,0}$ is the regularized incomplete gamma kernel introduced by \cite{emery2022gauss}:
\begin{equation*} 
    {\cal I}_{a,\alpha,d,0}(h) = 1- \frac{1}{\Gamma(\alpha-\frac{d}{2})} \Gamma^-\left(\alpha-\frac{d}{2},\frac{h^2}{a^2}\right) =  \frac{1}{\Gamma(\alpha-\frac{d}{2})} \Gamma^+\left(\alpha-\frac{d}{2},\frac{h^2}{a^2}\right), \quad h \geq 0.
\end{equation*}

Particular cases, which all belong to $\Phi_{\infty}$ insofar as the expression of these kernels does not depend on $d$, include:
\begin{enumerate}
\item[1.] The Gaussian kernel when $\alpha=\frac{d}{2}+1$ \citep[8.4.8]{Olver}:
\begin{equation*} 
    {\cal I}_{a,\frac{d}{2}+1,d,0}(h) = {\cal G}_a(h), \quad h \geq 0.
\end{equation*}
\item[2.] A Gaussian-polynomial kernel when $\alpha=\frac{d}{2}+1+q$ with $q \in \mathbb{N}$ \citep[8.4.8]{Olver}:
\begin{equation*} 
    {\cal I}_{a,\frac{d}{2}+1+q,d,0}(h) = \exp\left(-\frac{h^2}{a^2}\right) \sum_{n=0}^q \frac{1}{n!} \left(\frac{h^2}{a^2}\right)^n, \quad h \geq 0.
\end{equation*}
\item[3.] The complementary error function when $\alpha=\frac{d+1}{2}$ \citep[8.4.6]{Olver}:
\begin{equation*} 
    {\cal I}_{a,\frac{d+1}{2},d,0}(h) = \text{erfc} \left(\frac{h}{a}\right), \quad h \geq 0.
\end{equation*}
This kernel has been introduced by \cite{dalenius} for $d=2$, and by \cite{gneiting1999radial} for any $d \in \mathbb{N}_{\geq 1}$. The latter author proved that the Euclid's hat ${\cal H}_{\btheta}$ with $\btheta = (a(\frac{d-1}{2})^{1/2},\frac{d+1}{2},\frac{d}{2}+1,d+1,d,0)^\top$ uniformly converges on $[0,+\infty)$ to ${\cal I}_{a,\frac{d+1}{2},d,0}$ as $d$ tends to infinity.
\end{enumerate}

\section{Sobolev Consequences under the Class ${\cal H}$} \label{sec: consequences}
Theorem 3 has implications in many branches of statistics, machine learning, and approximation theory. We describe some of them. 

\begin{enumerate}
\item[1.] Best linear unbiased prediction under a misspecified covariance kernel is an important subject in spatial statistics \citep{stein1, stein2} and approximation theory \citep{scheuerer2010regularity}. Typically, the performance of the kriging predictor under an incorrect class of covariance kernels is measured by comparison with the {\em true} kernel under fixed domain asymptotics, that is, considering observations that increase over a compact set in such a way that the distance between the observations tends to zero. \cite{stein1} proves that the equivalence of Gaussian measures is a sufficient condition to ensure asymptotic optimality of the kriging predictor under a misspecified covariance kernel. The Sobolev properties of the kernel are of crucial importance and have been used under this framework for the Mat\'ern class \citep{zhang2004} as well as for the Generalized Wendland class \citep{BFFP}. This work fixes the basis to understand optimal unbiased linear prediction for a wealth of kernels that have not been studied so far under this perspective. 

\item[2.] The \emph{screening effect} is also a well known problem in spatial statistics. It is used to describe a situation where the interpolant depends mostly on those observations that are located nearest to the predictand \citep{stein1}. Such a problem has been of interest to geostatisticians for decades \citep{Chiles2012} because it translates into the optimality property of reducing considerably the computational burden associated with the kriging predictor when handling large data sets. 
Quantifying screening effects under a specified class of kernels is a major task that relies on several aspects, {\em e.g.} the spatial design (how to locate the observation points), the dimension of the Euclidean space where the spatial domain is embedded, the covariance kernel attached to a Gaussian spatial random field (or, equivalently, its spectral density) and the mean-square differentiability in all directions of the spatial random field. We first note that the screening effect is often quantified in a very practical way \citep{Chiles2012}. A formalization of the same is due to \cite{stein-book}, who provided sufficient conditions for the screening effect to happen under a regular sampling design. 
\cite{stein2} conjectured that, under the spectral condition
\begin{equation} 
\label{condition1}
\lim_{\|\boldsymbol{\omega}\| \to \infty } \sup_{\| \boldsymbol{\tau}\|<R} \bigg | \frac{\widehat{K}(\boldsymbol{\omega}+\boldsymbol{\tau} )}{\widehat{K}(\boldsymbol{\omega})}  -1 \bigg |=0,
\end{equation} 
one has screening effect under an irregular asymptotic design, and showed that (\ref{condition1}) is verified for $d \leq 2$ with mean-square continuous but nondifferentiable random fields, under some specific designs. 
\cite{porcu2020stein} proved that (\ref{condition1}) holds for the Generalized Wendland kernel, the argument being based on the tails of the related spectrum, and hence on the Sobolev properties. By following this argument, it becomes straightforward to deduce that such a condition is verified for the class ${\cal H}_{\btheta}$ as well. 

\item[3.] Theoretical results related to Gaussian regression in machine learning %\citep{rasmussen2006gaussian} 
are strongly connected to the Sobolev properties, and we refer the reader to \cite{korte2023smoothness}. For example, Sobolev smoothness is of crucial importance in Bayesian contraction rates \citep{van2011information}. Similar results where Sobolev rates pop up are contained in \cite{schaback2006kernel}, \cite{scheuerer2013interpolation} and \cite{narcowich2006sobolev}. As mentioned in \cite{korte2023smoothness} (see the references therein), the Sobolev properties turn to be fundamental within uncertainty quantification in nonparametric methods. The popular maximum mean discrepancies \citep{oates2017control} have been coupled with kernel methods, for which Sobolev properties cover a fundamental role, and the reader is referred to the most recent contribution in this direction by \cite{barp2022riemann}. 
The class of kernels proposed in this paper is a very good candidate in all these directions. Additionally, the property of compact support allows for considerable computational gains while preserving the required smoothness properties. Hence, extension of the previously mentioned directions to this class becomes imperative. It is not clear to the authors how the {\em hole effect} will play (in any) role within these research direction, although this aspect deserves  attention. 

\item[4.] Kernel methods have been widely used in the last decade to solve some systems of partial differential equations (PDEs) that were originally proposed within the approximation theory framework by \citet{fasshauer1996solving}, and for which a Bayesian turnaround was provided by \citet{cockayne2019bayesian}. 
The very interesting connection between statistics and approximation theory is provided by two facts: (a) the conditional mean of the process constructed by \cite{cockayne2019bayesian} coincides with the \emph{symmetric collocation} method introduced by  \citet{fasshauer1996solving}, and (b) the conditional variance is actually a measure of uncertainty quantification for the solution, while allowing for a finite computational budget. Sobolev methods become important because implementation of these methods requires regularity of a given order for the paths of the associated Gaussian field. Oversmoothness would affect accuracy in uncertainty quantification. The ability to customize smoothness has normally been attributed to the Mat\'ern class, which satisfies a specific class of stochastic partial differential equations (SPDEs) %\citep{whittle} 
and has become especially popular within the SPDE approximation thanks to the masterpiece of  \cite{Lindgren:Rue:Lindstrom:2011} and subsequently \cite{bolin} and \cite{bolin2020rational}. The class ${\cal H}$ opens for a wide spectrum analysis of several classes of (S)PDEs in concert with their application to Bayesian computation as a probabilistic extension to meshless methods. 

\item[5.] The Bayesian community has been increasingly concerned with a specific class of PDE called \emph{Stein equation}, which is used to compute the posterior expectation of a given function. Numerical solutions of this equation involve kernel methods \citep{oates2017control,south2022semi}. Again, there is no surprise that smoothness plays a major role. Using the ${\cal H}$ class as a surrogate to the Mat\'ern kernel would allow, within this context, for computationally cheaper solutions while preserving the customizable smoothness property.

\end{enumerate}

\section{Concluding remarks}
\label{sec:concl}

The introduced kernel is flexible as it allows to model important features such as local properties (Sobolev spaces, fractal dimension, mean square differentiability) as well as compact or global supports, and negative correlations. The kernel attains a wealth of well known kernels as special cases, and as a by-product allows to attain local properties of previously proposed kernels that have not been proposed beforehand.  

This work has impact in all the areas of machine learning, statistics, numerical analysis, and approximation theory. We expect these communities to be largely engaged to explore further properties that are notoriously of crucial importance in specific disciplines. In machine learning, this kernel might be taken as a benchmark to study its properties in terms of maximum mean discrepancies under Stein kernel methods. Another important subject within both machine learning and statistics is to explore the properties of this kernel in terms of posterior contraction rates in Gaussian regression \citep{rosa2024posterior}. In spatial statistics, understanding the properties of the ${\cal H}$ class in terms of equivalence of Gaussian measures will have crucial importance as it will allow to understand the interpolation properties (kriging) under a wealth of kernels, generalizing considerably the works of \cite{zhang2004} and \cite{BFFP}.
 
Data Science is providing many challenges, including {\em fancy} data domains. It will be challenging to have similar constructions to ${\cal H}$ for non-Euclidean domains and under different metrics.

% Acknowledgements and Disclosure of Funding should go at the end, before appendices and references

\acks{This work was funded by the National Agency for Research and Development of Chile, under grants ANID FONDECYT 1210050 (X. Emery), ANID PIA AFB230001 (X. Emery), ANID FONDECYT 1240308  (M. Bevilacqua), ANID project Data Observatory Foundation DO210001 (M. Bevilacqua), and MATH-AMSUD1167 22-MATH-06 (AMSUD220041) (M. Bevilacqua). Emilio Porcu is indebted to Dr. Horst Simon for the insightful discussions related to unified Native Spaces.}


%\newpage

\appendix

\section{A Turning Bands Construction of the Generalized Hypergeometric Kernel}
\label{walk}

The turning bands operator \citep{Matheron:1973} transforms an isotropic kernel defined in $\mathbb{R}^{d+p}$ to another isotropic kernel defined in $\mathbb{R}^{d}$, with $p \in \mathbb{N}$. Specifically, let $C_{d+p} \in \Phi_{d+p}$ such that $C_{d+p}(\| \cdot \|_{d+p})$ is absolutely integrable in $\R^{d+p}$, and let $g_{d+p}$ be the associated $(d+p)$-radial Schoenberg density. The turning bands operator of order $p$ of $C_{d+p}$ is the mapping $C_{d} \in \Phi_{d}$ with $d$-radial Schoenberg density $g_{d+p}$. Accounting for (\ref{schoenberg2fourier}), this implies \citep[5.3 and 5.4'']{matheron1972} 
\begin{equation}
\label{turning1}
  C_{d+p}(h) = \frac{2 \Gamma(\frac{d+p}{2})}{ \Gamma(\frac{p}{2}) \Gamma(\frac{d}{2})} h^{2-d-p} \int_0^{h} u^{d-1} (h^2-u^2)^{\frac{p}{2}-1} C_{d}(u) {\rm d}u, \quad h > 0,
\end{equation}
and
\begin{equation}
\label{turning2}
    \widehat{C}_{d}(u) = \frac{\pi^{\frac{p}{2}}\Gamma(\frac{d}{2})}{\Gamma(\frac{d+p}{2})} u^{p} \widehat{C}_{d+p}(u), \quad u > 0,
\end{equation}
where $\widehat{C}_{d+p}$ and $\widehat{C}_{d}$ are the $(d+p)$- and $d$-radial spectral densities of $C_{d+p}$ and $C_{d}$, respectively.

When $p=2$, one has
\begin{equation}
\label{turning3}
      C_{d}(h) = \frac{h^{1-d}}{d} \frac{\partial [h^{d} C_{d+2}(h)]}{\partial h} , \quad h > 0.
\end{equation}

In the general case, when $p=2k$ with $k \in \mathbb{N}$, one can rewrite (\ref{turning1}) as
\begin{equation*}
  x^{\frac{d}{2}+k-1} C_{d+2k}(\sqrt{x}) = \frac{2 \Gamma(\frac{d}{2}+k)}{ \Gamma(k) \Gamma(\frac{d}{2})} \int_0^{\sqrt{x}} u^{d-1} (x-u^2)^{k-1} C_{d}(u) {\rm d}u
\end{equation*}
and differentiate $k$ times using \citet[0.410, 0.42 and 0.433.1]{grad} to obtain \citep[Supplementary Material, Lemma 2]{emery2025}
\begin{equation}
\label{explicitTB}
  C_{d}(h) =\sum_{q=0}^k \sum_{r=0}^{\max\{0,q-1\}} \frac{(-1)^r (k-q+1)_q (q)_r (q-r)_r \, h^{q-r} \, C_{d+2k}^{(q-r)}(h)}{2^{q+r} q! \, r! (\frac{d}{2})_q}.
\end{equation}

Also, $C_{d+2k}$ and $C_{d}$ have the same value at the origin: this stems from (\ref{schoenbergmeasure}) and the fact that both kernels have, by definition, the same Schoenberg measure. Furthermore, if $C_{d+2k}$ is continuous, nonnegative and supported in $[0,a]$, the following properties are a consequence of (\ref{turning3}) and (\ref{explicitTB}):
\begin{enumerate}
\item[1.] $C_{d}$ vanish for $h > a$ because so does $C_{d+2k}$: both kernels are compactly supported.
\item[2.] If $C_{d+2k}(0)-C_{d+2k}$ is, up to a positive constant, equivalent to $h \mapsto h^{\eta}$ (with $\eta \in \mathbb{R}_{>0}$) as $h \to 0^+$, then so does $C_{d}(0)-C_{d}$: the two kernels have the same smoothness.
\item[3.] $C_{d}$ has at least $k$ different zeros in $(0,a)$, which results from (\ref{turning3}) and a recursive application of Rolle's theorem: hole effects appear when $k>0$.
\end{enumerate}

In particular, let $\btheta = (a,\alpha,\beta,\gamma,d,k)^{\top}$ satisfying conditions (A.1) to (A.4). Owing to (\ref{hygeodensity}) and (\ref{turning2}), it is seen that ${\cal H}_{\btheta}$ is obtained, up to a positive factor, by applying the turning bands operator of order $2k$ to ${\cal H}_{\btheta^\prime}$ with $\btheta^\prime = (a,\alpha,\beta,\gamma,d+2k,0)^{\top}$ that also satisfies conditions (A.1) to (A.4). This construction generalizes that of \cite{gneiting2002compactly}, who applied the turning bands operator to a subfamily of ordinary Wendland kernels (the case $\alpha=\frac{d+3}{2}$, $\beta \geq \frac{3\alpha}{2}$, $\gamma=\beta+\frac{1}{2}$ and $k=1$, which leads to a subcase of the hole effect Generalized Wendland kernel presented in Proposition 4).

Since ${\cal H}_{\btheta^\prime}$ is continuous, nonnegative and supported in $[0,a]$ \citep{emery2022gauss}, the aforementioned properties hold. In particular, ${\cal H}_{\btheta}$ has the same smoothness as ${\cal H}_{\btheta^\prime}$ and exhibits one or more hole effects as soon as $k>0$.

\section{Analytical Expressions of the Generalized Hypergeometric Kernel}
\label{alternativesGH}

\noindent \textbf{Expressions in terms of Gauss hypergeometric functions.} 
Using formulae 7.4.1.2 of \citet[7.4.1.2]{prud} and 5.5.3 of \cite{Olver}, one can rewrite (\ref{GeneralizedHypergeometric}) in terms of Gauss hypergeometric functions:
\begin{equation}
\label{GeneralizedHypergeometric2}
\begin{split}
    {\cal H}_{\btheta}(h)  &= \sum_{n=0}^k \frac{(-1)^n k! (1+\frac{d}{2}+k-\beta)_n (1+\frac{d}{2}+k-\gamma)_n}{n! (k-n)! (1-\frac{d}{2}-n)_n (1+\frac{d}{2}+k-\alpha)_n} \\
    &\qquad \times \left(\frac{h}{a}\right)^{2n} \textstyle\pFq{2}{1}{1+\frac{d}{2}+k-\beta+n,1+\frac{d}{2}+k-\gamma+n}{1+\frac{d}{2}+k-\alpha+n}{\frac{h^2}{a^2}}\\
    &+ \frac{\Gamma(\beta-\frac{d}{2}-k) \Gamma(\gamma-\frac{d}{2}-k) \Gamma(\frac{d}{2})\Gamma(\frac{d}{2}+k-\alpha)}{\Gamma(\frac{d}{2}+k) \Gamma(\alpha-\frac{d}{2}-k)\Gamma(\beta-\alpha)\Gamma(\gamma-\alpha)}\\
    &\qquad \times \sum_{n=0}^k \frac{(-1)^{n+k} k! (1-\alpha)_{k-n} (1+\alpha-\beta)_n (1+\alpha-\gamma)_n}{n! (k-n)! (1+\alpha-\frac{d}{2}-k)_n}\\
    &\qquad \times \left(\frac{h}{a}\right)^{2\alpha-d-2k+2n} \textstyle\pFq{2}{1}{1+\alpha-\beta+n,1+\alpha-\gamma+n}{1+\alpha-\frac{d}{2}-k+n}{\frac{h^2}{a^2}}, \quad 0 \leq h < a.
\end{split}
\end{equation}

An alternative is to apply (\ref{explicitTB}) to the Gauss hypergeometric kernel ${\cal H}_{\btheta^\prime}$ as defined in Appendix \ref{walk}. Using (\ref{hygeo2F1}) with $d+2k$ instead of $d$, \citet[0.432]{grad} and \citet[15.5.4]{Olver}, one finds:
\begin{equation}
\label{GeneralizedHypergeometric4}
\begin{split}
    {\cal H}_{\btheta}(h) &=\sum_{q=0}^k \sum_{r=0}^q \sum_{s=0}^{\max \{0,q-r-1\}} \frac{(-1)^q (k-q+1)_q (q)_r (q-r)_r (q-r-2s+1)_{2s}}{2^{2r+2s} q! \, r! \, s! \, (\frac{d}{2})_q} \\ 
    &\quad \times \frac{\Gamma(\beta-\frac{d}{2}-k)\Gamma(\gamma-\frac{d}{2}-k) }{\Gamma(\beta-\alpha+\gamma-\frac{d}{2}-k-q+r+s) \Gamma(\alpha-\frac{d}{2}-k)} \\
    &\quad \times \left(\frac{h}{a}\right)^{q-r} \left(1-\frac{h}{a}\right)_+^{\beta-\alpha+\gamma-\frac{d}{2}-k-1-s} \left(1+\frac{h}{a}\right)^{\beta-\alpha+\gamma-\frac{d}{2}-k-q+r+s-1} \\
    &\quad \times \textstyle\pFq{2}{1}{\beta-\alpha,\gamma-\alpha}{\beta-\alpha+\gamma-\frac{d}{2}-k-q+r+s}{1-\frac{h^2}{a^2}}, \quad 0 < h \leq a.  
\end{split}
\end{equation}

The right-hand side of (\ref{GeneralizedHypergeometric4}) is a continuous function on $(0,a]$ that vanishes at $h=a$ under conditions (A.1) to (A.3), even if condition (A.4) does not hold, which proves that ${\cal H}_{\btheta}$ can be defined by continuation when $\alpha-\frac{d}{2}$ is an integer. This continuation is still a member of $\Phi_d$, insofar as it is the image by the turning bands operator of order $2k$ of a function (${\cal H}_{\btheta^\prime}$) belonging to $\Phi_{d+2k}$.\\

\noindent \textbf{Expression in terms of a Meijer function.} 
Using formulae 8.2.2.3 and 8.2.2.15 of \cite{prud}, one can rewrite (\ref{GeneralizedHypergeometric}) in terms of a Meijer-$G$ function:
\begin{equation}
    \label{GeneralizedHypergeometric3}
    \begin{split}
    {\cal H}_{\btheta}(h) = \begin{cases}
    0 \text{ if $h \geq a$}\\
    \frac{\Gamma(\frac{d}{2}) \Gamma(\beta-\frac{d}{2}-k) \Gamma(\gamma-\frac{d}{2}-k)}{\Gamma(\alpha-\frac{d}{2}-k) \Gamma(\frac{d}{2}+k)}  \textstyle\G{2,1}{3,3}{1-\frac{d}{2}-k,\beta-\frac{d}{2}-k,\gamma-\frac{d}{2}-k}{0,\alpha-\frac{d}{2}-k,1-\frac{d}{2}}{\frac{h^2}{a^2}} \text{ if $0<h<a$}\\
    1 \text{ if $h = 0$}.\end{cases}
    \end{split}
\end{equation}

One can also study the behavior of ${\cal H}_{\btheta}$ near the range by using the expansion of the Meijer-$G$ function of argument close to $1$ \citep[8.2.2.60]{prud}. It comes:
\begin{equation}
{\cal H}_{\btheta}(h) \underset{h \to a^-}{\sim} \varsigma \left(1-\frac{h}{a}\right)^{\beta+\gamma-\alpha-2k-\frac{d}{2}-1}
\end{equation}
with $\varsigma \neq 0$, provided conditions (A.1) to (A.4) hold, $\beta + \gamma \notin \mathbb{N}$ and $\beta + \gamma - \alpha - \frac{d}{2}\notin \mathbb{N}$. In this setting, ${\cal H}_{\btheta}$ has left derivatives of orders $1$ to $p$ that vanish at $h=a$ (hence, ${\cal H}_{\btheta}$ is $p$-times differentiable at $h=a$) if, and only if, $\beta+\gamma-\alpha-2k-\frac{d}{2}-1>p$. By continuation, ${\cal H}_{\btheta}$ remains $p$-times differentiable at $h=a$ when either $\beta + \gamma \in \mathbb{N}$ or $\beta + \gamma - \alpha - \frac{d}{2} \in \mathbb{N}$. 

\section{Proofs}
\label{appProofs}

\noindent \textbf{Lemma 1} \citep[p. 81]{Polya-Szego:1998}
%\label{Dini}
{\it Let $\{ f_n : n \in \mathbb{N} \}$ be a sequence of real-valued non-increasing functions on $[0,b]$, with $b \in (0,+\infty]$, that converge pointwise to a continuous function $f$ on $[0,b]$. Then, the convergence is uniform on $[0,b]$.}
\hfill\BlackBox \\

\noindent \textbf{Proof of Theorems 1 and 2}
Let $a, \alpha, \beta, \gamma, \tau \in \mathbb{R}_{> 0}$, $d \in \mathbb{N}_{\geq 1}$, and $\kappa \geq 0$ such that  $2(\beta-\alpha) (\gamma-\alpha) \geq \alpha$, $2(\beta+\gamma) \geq 6 \alpha +1$ and $\alpha-\frac{d}{2}-\kappa \notin \mathbb{N}$. Let $\btheta = (a,\alpha,\beta,\gamma,d,\kappa)^{\top}$ and define the mapping ${\widehat{{\cal H}}}_{\btheta}$ through
\begin{equation*}
%\label{spectraldensity}
    \widehat{{\cal H}}_{\btheta}(u) = \tau u^{2\kappa} 
    \textstyle\pFq{1}{2}{\alpha}{\beta,\gamma}{-\frac{ a^2 u^2}{4}}, \quad u \in [0,+\infty),
\end{equation*}
which is nonnegative on $[0,+\infty)$ owing to Theorem 4.2 in \cite{cho2020rational}. By expressing the Bessel function $J$ in terms of the generalized hypergeometric function ${}_0F_1$ \citep[10.16.9]{Olver} and using formulae 5.1 in \cite{Miller1998} (valid under the additional condition $\alpha > \frac{d+1}{4}+\kappa$), the Fourier-Hankel transform (\ref{fourier1}) of $\widehat{{\cal H}}_{\btheta}$ at $h>0$ is found to be
\begin{equation*}
    \begin{split}
    {\cal H}_{\btheta}(h) &= \frac{2 \tau \pi^{\frac{d}{2}}}{\Gamma(\frac{d}{2})} \int_0^{+\infty} u^{d+2\kappa-1}\, 
    \textstyle\pFq{0}{1}{-}{\frac{d}{2}}{-\frac{u^2 h^2}{4}}
    \textstyle\pFq{1}{2}{\alpha}{\beta,\gamma}{-\frac{ a^2 u^2}{4}}
    {\rm d}u\\
    &= \begin{cases}
    \frac{2 \tau \pi^{\frac{d}{2}}}{\Gamma(\frac{d}{2})} \frac{\Gamma(\frac{d}{2}+\kappa) \Gamma(\frac{d}{2})}{2 (h/2)^{d+2\kappa} \Gamma(-\kappa)} \, \textstyle\pFq{3}{2}{\alpha,\frac{d}{2}+\kappa,\kappa+1}{\beta,\gamma}{\frac{a^2}{h^2}} \text{ if $a < h$}\\
    \frac{\tau \pi^{\frac{d}{2}}}{\Gamma(\frac{d}{2})} \frac{2^{d+2\kappa}\Gamma(\frac{d}{2}+\kappa) \Gamma(\alpha-\frac{d}{2}-\kappa) \Gamma(\beta) \Gamma(\gamma)}{a^{d+2\kappa}\Gamma(\alpha) \Gamma(\beta-\frac{d}{2}-\kappa) \Gamma(\gamma-\frac{d}{2}-\kappa)} \textstyle\pFq{3}{2}{\frac{d}{2}+\kappa,1+\frac{d}{2}+\kappa-\beta,1+\frac{d}{2}+\kappa-\gamma}{1+\frac{d}{2}+\kappa-\alpha,\frac{d}{2}}{\frac{h^2}{a^2}} \\
    + \frac{\tau \pi^{\frac{d}{2}}}{\Gamma(\frac{d}{2})} \frac{2^{d+2\kappa}\Gamma(\frac{d}{2})\Gamma(\beta)\Gamma(\gamma)\Gamma(\frac{d}{2}+\kappa-\alpha)}{a^{2\alpha}\Gamma(\beta-\alpha)\Gamma(\gamma-\alpha) \Gamma(\alpha-\kappa)} h^{2\alpha-d-2\kappa} \, \textstyle\pFq{3}{2}{\alpha,1+\alpha-\beta,1+\alpha-\gamma}{1+\alpha-\frac{d}{2}-\kappa,\alpha-\kappa}{\frac{h^2}{a^2}}  \text{ if $0 < h < a,$} 
    \end{cases}
    \end{split}
\end{equation*}
which is well-defined for $h \in (0,a) \cup (a,+\infty)$ and extendable at $h=a$ by continuity under the conditions stated above. Now, if $\kappa \in \mathbb{N}$, ${\cal H}_{\btheta}$ turns out to be identically zero on $(a,+\infty)$, and if $\alpha > \frac{d}{2}+\kappa$, it can be extended by continuity at $h=0$; in such a case, to obtain ${\cal H}_{\btheta}(0)=1$, we have to set
\begin{equation*}
    \tau = \frac{a^{d+2k}\Gamma(\frac{d}{2})\Gamma(\alpha) \Gamma(\beta-\frac{d}{2}-\kappa) \Gamma(\gamma-\frac{d}{2}-\kappa)}{\pi^{\frac{d}{2}} 2^{d+2\kappa}\Gamma(\frac{d}{2}+\kappa) \Gamma(\alpha-\frac{d}{2}-\kappa) \Gamma(\beta) \Gamma(\gamma)},
\end{equation*}
which yields the announced kernel ${\cal H}_{\btheta}$ (Theorem 1) and $d$-radial spectral density $\widehat{{\cal H}}_{\btheta}$ (Theorem 2). The previous arguments also imply that ${\cal H}_{\btheta}$ is continuous on $[0,+\infty)$.\\

\noindent \textbf{Proof of Theorem 3}
Results in \cite{cho2020rational} show that the function $\widehat{{\cal H}}_{\btheta}$ in (\ref{hygeodensity}) is strictly positive if the following conditions hold:
\begin{itemize}
    \item[(B.1)] $\alpha > 0$;
    \item[(B.2)] $2(\beta-\alpha) (\gamma-\alpha) \geq \alpha$;
    \item[(B.3)] $2(\beta+\gamma) > 6 \alpha +1$.    
\end{itemize}
These conditions are met when conditions (A.1) to (A.3) hold and $2(\beta+\gamma) \neq 6 \alpha +1$.

Additionally, the mapping $\|\cdot\|_d \mapsto \widehat{H}_{\btheta}(\|\cdot\|_d)$ is absolutely integrable in $\R^d$ under conditions (A.1) to (A.3). To prove that condition (\ref{bounded}) holds for some $0<c_1< c_2<\infty$ and some $s>d/2$, we invoke the asymptotic expansion for the generalized hypergeometric function ${}_1F_2$ \citep[p. 146]{mathai}:
\begin{equation*}
\textstyle\pFq{1}{2}{\alpha}{\beta,\gamma}{-\frac{x^2}{4}} \underset{x \to +\infty}{\sim} A x^{\alpha-\beta-\gamma+\frac{1}{2}} \cos(x+B) + C x^{-2\alpha},
\end{equation*}
with $A, B, C$ being real values. As $\alpha-\beta-\gamma+ \frac{1}{2} < -2\alpha $ due to condition (B.3), the leading term is the last term in $x^{-2\alpha}$. Accordingly:
\begin{equation*}
\widehat{{\cal H}}_{\btheta}(u) \underset{u \to +\infty}{\sim} \zeta_{\btheta} \; u^{2k-2\alpha}
\end{equation*}
with $\zeta_{\btheta} \in \mathbb{R}_{>0}$ given that $\widehat{{\cal H}}_{\btheta}$ is positive.

\noindent \textbf{Proof of Proposition 1}
The identity (\ref{hygeo2F1}) is obtained by use of (\ref{GeneralizedHypergeometric2}) and formula 15.8.4 in \cite{Olver}. 

\noindent \textbf{Proof of Proposition 2}
The identity (\ref{truncatedpow}) is obtained by use of (\ref{GeneralizedHypergeometric2}) and of the series expansion of the hypergeometric function ${}_2F_1$ \citep[15.2.1]{Olver}, being terminating series under the conditions stated in the proposition. 

\noindent \textbf{Proof of Proposition 3}
See \citet[eq. 15]{Chernih} and \cite{bevi2024} to establish (\ref{geneWendl}) and the equivalence between conditions (A.1) to (A.3) of Theorem 1 and the stated conditions $\xi > -\frac{1}{2}$ and $\nu \geq \nu_{\min}(\xi,d)$.

\noindent \textbf{Proof of Proposition 4}
See \cite{emery2025} to establish (\ref{wendlandext}) and the equivalence between conditions (A.1) to (A.3) of Theorem 1 and the stated conditions $k \in \mathbb{N}$, $\xi > -\frac{1}{2}$ and $\nu \geq \nu_{\min}(\xi,d+2k)$.

\noindent \textbf{Proof of Proposition 5}
The result stems from (\ref{hygeo2F1}) and formula 15.4.17 of \cite{Olver}.

\noindent \textbf{Proof of Proposition 6}
This is a particular case of Proposition 4 with $\xi=0$, see \cite{emery2025}.

\noindent \textbf{Proof of Propositions 7 and 8}
See \cite{emery2025}.

\noindent \textbf{Proof of Proposition 9} 
We just need to establish the convergence of ${\cal M}_{a,k + \frac{1}{2},d,k}$ to ${\cal J}_{a,d}$: the convergence of ${\cal W}_{\mu a,k,\mu,d,k}$ to ${\cal J}_{a,d}$ is then a consequence of Proposition 8. The proof relies on the following alternative expression of ${\cal M}_{a,\nu,d,k}$ \citep{emery2025}:
\begin{equation}
\label{holeeffectMatern}
    \begin{split}
    {\cal M}_{a,\nu,d,k}(h) &= \textstyle\pFq{1}{2}{k+\frac{d}{2}}{1-\nu,\frac{d}{2}}{\frac{h^2}{4a^2}} \\
    &+ \frac{\Gamma(\nu+\frac{d}{2}+k) \Gamma(\frac{d}{2}) \Gamma(-\nu)}{\Gamma(\frac{d}{2}+k) \Gamma(\nu)\Gamma(\nu+\frac{d}{2})} \left(\frac{h}{2a}\right)^{2\nu} \textstyle\pFq{1}{2}{\nu+\frac{d}{2}+k}{\nu+\frac{d}{2},\nu+1}{\frac{h^2}{4a^2}}, \quad h \geq 0, \nu \notin \mathbb{N}_{\geq 1}.
    \end{split}
\end{equation}

Let us write the series representation of the hypergeometric ${}_1F_2$ functions in (\ref{holeeffectMatern}). Concerning the first ${}_1F_2$ function, one has:
\begin{equation}
\label{generalizedMatern2}
\textstyle\pFq{1}{2}{k+\frac{d}{2}}{1-\nu,\frac{d}{2}}{\frac{h^2}{4a^2}} = \sum_{n=0}^{+\infty} \frac{(k+\frac{d}{2})_n}{(1-\nu)_n (\frac{d}{2})_n n!} \left(\frac{h^2}{4a^2}\right)^{n}, \quad h \geq 0.
\end{equation}

Let us choose $\nu = k + \frac{1}{2}$. As $k$ tends to infinity, $(k+\frac{d}{2})_n$ and $(1-\nu)_n$ are equivalent to $k^n$ and $(-k)^n$, respectively \citep[5.5.3 and 5.11.12]{Olver}. We can therefore split the alternating series (\ref{generalizedMatern2}) into the difference of two series with strictly positive terms, which tend to
$$\sum_{n=0}^{+\infty} \frac{1}{(\frac{d}{2})_{2n} ({2n})!} \left(\frac{h^2}{4 a^2}\right)^{2n} = \frac{1}{2} \left[\textstyle\pFq{0}{1}{-}{\frac{d}{2}}{\frac{h^2}{4a^2}} + \textstyle\pFq{0}{1}{-}{\frac{d}{2}}{-\frac{h^2}{4a^2}} \right]$$
and
$$\sum_{n=0}^{+\infty} \frac{1}{(\frac{d}{2})_{2n+1} ({2n+1})!} \left(\frac{h^2}{4 a^2}\right)^{2n+1} = \frac{1}{2} \left[\textstyle\pFq{0}{1}{-}{\frac{d}{2}}{\frac{h^2}{4a^2}}-\textstyle\pFq{0}{1}{-}{\frac{d}{2}}{-\frac{h^2}{4a^2}}\right],$$
respectively, with all the ${}_0F_1$ terms being nonzero for almost all $h \in [0,+\infty)$. In both cases, the convergence is uniform on $[0,+\infty)$ owing to Lemma 1.

Accordingly, if $\nu = k + \frac{1}{2}$ and $k$ tends to infinity ($d$, $h$ and $a$ fixed), one obtains \citep[10.16.9]{Olver}
\begin{equation*}
\textstyle\pFq{1}{2}{k+\frac{d}{2}}{\frac{1}{2}-k,\frac{d}{2}}{\frac{h^2}{4a^2}} \to \textstyle\pFq{0}{1}{-}{\frac{d}{2}}{-\frac{h^2}{4a^2}} = {\cal J}_{a,d}(h),
\end{equation*}
the convergence being uniform on $[0,+\infty)$.

Concerning the second hypergeometric function in (\ref{holeeffectMatern}), the series expansion only has positive terms and one obtains, under the same conditions on $k$ and $\nu$:
\begin{equation*}
\textstyle\pFq{1}{2}{\frac{d+1}{2}+2k}{k+\frac{d+1}{2},k+\frac{3}{2}}{\frac{h^2}{4a^2}} \to \textstyle\pFq{0}{0}{-}{-}{0} = 1,
\end{equation*}
with again the convergence being uniform on $[0,+\infty)$.

The result of the proposition follows from the fact that
$$h \mapsto \left(\frac{h}{2a}\right)^{2k+1} \frac{\Gamma(2k+\frac{d+1}{2}) \Gamma(\frac{d}{2}) \Gamma(-k-\frac{1}{2})}{\Gamma(k+\frac{d}{2}) \Gamma(k+\frac{1}{2})\Gamma(k+\frac{d+1}{2})}$$ uniformly tends to zero on any bounded interval of $[0,+\infty)$ as $k$ tends to $+\infty$.

\noindent \textbf{Proof of Proposition 10} 
\cite{emery2025} showed that, for $\nu \geq \frac{3}{2}$, $\mu \geq \max\{\nu+\frac{d}{2},\frac{1}{a}\}$ and $h \in (0,\mu a -1)$, one has:
\begin{equation*}
\begin{split}
    0 &\leq {\cal M}_{a,\nu}(h)- \frac{\Gamma(\mu+1) \mu^{2\nu-1} }{\Gamma(\mu+2\nu)} {\cal W}_{\mu a,\nu-\frac{1}{2},\mu,d,0}(h) 
    \leq \frac{4 \nu (\nu+1)}{\mu} + \frac{\Gamma(2\nu-1,\mu)}{\Gamma(2\nu-1)}.
\end{split}
\end{equation*}

Let $a = \frac{b}{2\sqrt{\nu}}$ with $b>0$ fixed. Also, let $\nu$ tend to infinity in such a way that $\mu \nu^{-2}$ tends to infinity. Then:
\begin{itemize}
\item $\frac{\Gamma(\mu+1) \mu^{2\nu-1} }{\Gamma(\mu+2\nu)} \to 1$ \citep[5.11.12]{Olver};
\item $\frac{4 \nu (\nu+1)}{\mu} + \frac{\Gamma(2\nu-1,\mu)}{\Gamma(2\nu-1)} \to 0$;
\item ${\cal M}_{a,\nu}(h) \to {\cal G}_{b}(h)$ for any $h \in [0,+\infty)$ \citep{stein-book}, with the convergence being uniform owing to Lemma 1. 
\end{itemize}

One deduces the uniform convergence of ${\cal W}_{\mu a,\nu-\frac{1}{2},\mu,d,0}$ to ${\cal G}_{b}$ on $[0,+\infty)$.

\noindent \textbf{Proof of Propositions 11 and 12} 
Arguments similar to those used in the proof of Proposition 9 allow establishing the following uniform convergences on any bounded interval $I$ of $[0,+\infty)$ as $\nu$ tends to infinity:
\begin{equation*}
\textstyle\pFq{1}{2}{\nu+\frac{d}{2}+k}{\nu+\frac{d}{2},\nu+1}{\frac{\nu h^2}{a^2}} \to \textstyle\pFq{0}{0}{-}{-}{\frac{h^2}{a^2}} = \exp\left(\frac{h^2}{a^2}\right) , \quad h \in I,
\end{equation*}
and
\begin{equation*}
\textstyle\pFq{1}{2}{k+\frac{d}{2}}{1-\nu,\frac{d}{2}}{\frac{\nu h^2}{a^2}} \to 
\textstyle\pFq{1}{1}{k+\frac{d}{2}}{\frac{d}{2}}{-\frac{h^2}{a^2}}, \quad h \in I,
\end{equation*}
with the latter expression matching the Gaussian kernel (\ref{gaulaguerre}) owing to formulae 7.11.1.8 of \cite{prud} and 5.5.3 of \cite{Olver}. Furthermore, based on formulae 5.5.3 and 5.11.7 of \cite{Olver}, one has the following uniform convergence on any bounded interval $I$ of $[0,+\infty)$ when $\nu$ is a half-integer tending to infinity:
\begin{equation*}
\frac{\Gamma(\nu+\frac{d}{2}+k) \Gamma(\frac{d}{2}) \Gamma(-\nu)}{\Gamma(k+\frac{d}{2}) \Gamma(\nu)\Gamma(\nu+\frac{d}{2})} \left(\frac{\nu h^2}{a^2}\right)^{\nu} \to 0, \quad h \in I.
\end{equation*}

Equation (\ref{holeeffectMatern}) implies the uniform convergence of ${\cal M}_{\sqrt{\nu} a,\nu,d,k}$ to ${\cal G}_{a,d,k}$ on $I$ as $\nu$ is a half-integer tending to infinity. The uniform convergence of ${\cal W}_{\mu \sqrt{\nu} a,\nu-\frac{1}{2},\mu,d,k}$ to the same kernel as $\frac{\mu}{\nu}$ also tends to infinity stems from Proposition 8. In passing, these convergences prove that ${\cal G}_{a,d,k}$ belongs to $\Phi_d$ as a continuous function that is the pointwise limit of members of $\Phi_d$.

\noindent \textbf{Proof of Proposition 13}
The proof stems from formula 8.1.8 of \cite{Szego} and formula 5.11.12 of \cite{Olver}.

\noindent \textbf{Proof of Proposition 14}
Again, the same argument as in the proof of Proposition 9 allows establishing the following uniform convergences on any bounded interval $I$ of $[0,+\infty)$ as $\gamma$ tends to infinity and $a/\sqrt{\gamma}$ tends to $b>0$:
\begin{equation} 
\label{incgamma1}
    \textstyle\pFq{3}{2}{\alpha,1+\alpha-\beta,1+\alpha-\gamma}{1+\alpha-\frac{d}{2}-k,\alpha-k}{{\frac{h^2}{a^2}}} \to \textstyle\pFq{2}{2}{\alpha,1+\alpha-\beta}{1+\alpha-\frac{d}{2}-k,\alpha-k}{-\frac{h^2}{b^2}}, \quad h \in I,
\end{equation}
\begin{equation*} 
    \textstyle\pFq{3}{2}{\frac{d}{2}+k,1+\frac{d}{2}+k-\beta,1+\frac{d}{2}+k-\gamma}{1+\frac{d}{2}+k-\alpha,\frac{d}{2}}{\frac{h^2}{a^2}} \to \textstyle\pFq{2}{2}{\frac{d}{2}+k,1+\frac{d}{2}+k-\beta}{1+\frac{d}{2}+k-\alpha,\frac{d}{2}}{-\frac{h^2}{b^2}}, \quad h \in I,
\end{equation*}
and
\begin{equation*} 
    \frac{\Gamma(\gamma-\frac{d}{2}-k)}{\Gamma(\gamma-\alpha)} \left ( \frac{h}{a} \right )^{2\alpha-d-2k} \to \left ( \frac{h}{b} \right )^{2\alpha-d-2k}, \quad h \in I.
\end{equation*}

Accordingly, as $\gamma$ tends to infinity, the kernel ${\cal H}_{\btheta}$ as defined in (\ref{GeneralizedHypergeometric}) with $a=b\sqrt{\gamma}$ and $\beta=1+\frac{d}{2}+k$ uniformly converges on $I$ to the kernel ${\cal I}_{b,\alpha,d,k}$ defined by
\begin{equation*} 
\begin{split}    
    {\cal I}_{b,\alpha,d,k}(h) &= 1-\frac{(\alpha-k)_k}{(\frac{d}{2})_k \Gamma(\alpha-\frac{d}{2}-k+1)}  
    \left ( \frac{h}{b} \right )^{2\alpha-d-2k} \textstyle\pFq{2}{2}{\alpha,\alpha-\frac{d}{2}-k}{1+\alpha-\frac{d}{2}-k,\alpha-k}{-\frac{h^2}{b^2}} \\
    &= 1-\sum_{n=0}^k \frac{(-1)^n k! (\alpha-k)_k (\alpha-\frac{d}{2}-k)_n}{n! (k-n)! \Gamma(1+\alpha-\frac{d}{2}-k+n) \, (\alpha-k)_n (\frac{d}{2})_k}  
     \\
    & \quad \times \left ( \frac{h}{b} \right )^{2\alpha-d-2k+2n} \textstyle\pFq{1}{1}{\alpha-\frac{d}{2}-k+n}{1+\alpha-\frac{d}{2}-k+n}{-\frac{h^2}{b^2}},
\end{split}
\end{equation*}
where we accounted for the fact that the right-hand side of (\ref{incgamma1}) is identically equal to $1$, for Theorem 2.1 of \cite{withers} to expand the ${}_2F_2$ function into a series of ${}_1F_1$ functions. The claim of the proposition follows from \citet[8.5.1]{Olver}.

\vskip 0.2in
\bibliography{JMLR}

\end{document}